\documentclass{article}

\usepackage{microtype}
\usepackage{graphicx}
\usepackage{subcaption}
\usepackage{booktabs} 

\usepackage{hyperref}
\usepackage{multirow}
\usepackage{xurl} 
\usepackage{tabularx}
\usepackage{booktabs}
\usepackage{pifont}   
\usepackage{makecell}

\usepackage[preprint]{icml2026}

\usepackage{amsmath}
\usepackage{amssymb}
\usepackage{mathtools}
\usepackage{amsthm}
\usepackage{bbding}

\usepackage[capitalize,noabbrev]{cleveref}

\theoremstyle{plain}

\theoremstyle{definition}

\theoremstyle{remark}

\newcommand{\xname}{ChipBench}

\usepackage[textsize=tiny]{todonotes}
\usepackage{longtable}
\usepackage{booktabs} 
\usepackage{graphicx} 
\usepackage{xcolor}
\usepackage{tcolorbox}
\usepackage{colortbl}
\tcbuselibrary{listings, skins, raster}

\begin{document}

\twocolumn[
  \icmltitle{\xname: A Next-Step Benchmark for Evaluating LLM Performance in AI-Aided Chip Design}



  \icmlsetsymbol{equal}{*}

  \begin{icmlauthorlist}
    \icmlauthor{Zhongkai Yu}{equal,ucsd}
    \icmlauthor{Chenyang Zhou}{equal,aaa}
    \icmlauthor{Yichen Lin}{ucsd}
    \icmlauthor{Hejia Zhang}{ucsd}
    \icmlauthor{Haotian Ye}{ucsd}
    \icmlauthor{Junxia Cui}{ucsd}
    \icmlauthor{Zaifeng Pan}{ucsd}
    \icmlauthor{Jishen Zhao}{ucsd}
    \icmlauthor{Yufei Ding}{ucsd}
  \end{icmlauthorlist}

  \icmlaffiliation{ucsd}{Department of Computer Science and Engineering, University of California San Diego, La Jolla, US}
  \icmlaffiliation{aaa}{Columbia University, New York, US}

  \icmlcorrespondingauthor{Firstname1 Lastname1}{first1.last1@xxx.edu}

  \icmlkeywords{Machine Learning, ICML}

  \vskip 0.3in
]



\printAffiliationsAndNotice{\icmlEqualContribution}

\begin{abstract}
While Large Language Models (LLMs) show significant potential in hardware engineering, current benchmarks suffer from saturation and limited task diversity, failing to reflect LLMs' performance in real industrial workflows.
To address this gap, we propose a comprehensive benchmark for AI-aided chip design that rigorously evaluates LLMs across three critical tasks: Verilog generation, debugging, and reference model generation. 
Our benchmark features 44 realistic modules with complex hierarchical structures, 89 systematic debugging cases, and 132 reference model samples across Python, SystemC, and CXXRTL.
Evaluation results reveal substantial performance gaps, with state-of-the-art Claude-4.5-opus achieving only 30.74\% on Verilog generation and 13.33\% on Python reference model generation, demonstrating significant challenges compared to existing saturated benchmarks where SOTA models achieve over 95\% pass rates.
Additionally, to help enhance LLM reference model generation, we provide an automated toolbox for high-quality training data generation, facilitating future research in this underexplored domain.
Our code is available at \href{https://github.com/zhongkaiyu/ChipBench.git}{https://github.com/zhongkaiyu/ChipBench.git}.


\end{abstract}

\section{Introduction}
Large Language Models (LLMs) have demonstrated remarkable capabilities in 
conversational AI and code generation~\cite{cai2024advancing, du2024evaluating, 
gu2025effectiveness, li2025large, liu2023your, nadimi2024multi, ross2023programmer, 
thakur2024verigen, wang2023codet5+, zhong2024can}. The integration of agentic 
techniques have further amplified these abilities, with systems like Claude and 
Cursor excelling in high-level languages such as Python and C++~\cite{wang2025ai, 
guo2025comprehensive, ishibashi2024self, islam2024mapcoder, dong2025survey}. 
Recently, this success has extended to the semiconductor domain, where LLMs show promise in chip design~\cite{gai2025exploring, abdollahi2024hardware, swaroopa2024evaluating, hu2024uvllm, he2025large, firouzi2025chipmnd, 
ho2025verilogcoder, yu2025spec2rtl, thakur2024verigen}.

To evaluate LLM capabilities in chip design, benchmarks like VerilogEval~\cite{liu2023verilogeval, pinckney2025revisiting} and RTLLM~\cite{lu2024rtllm, liu2024openllm} have been proposed. However, these benchmarks have become inadequate due to rapid LLM advancement. State-of-the-art LLM-based Verilog generators, such as MAGE~\cite{zhao2025mage}, have already achieved over 95\% accuracy on these benchmarks, indicating saturation and highlighting the need for more challenging and comprehensive evaluations. 
We identify three critical limitations of existing benchmarks that hinder accurate assessment of LLM capabilities for industrial deployment:

\begin{figure*}[t!]
    \centering
    \includegraphics[width=0.98\linewidth]{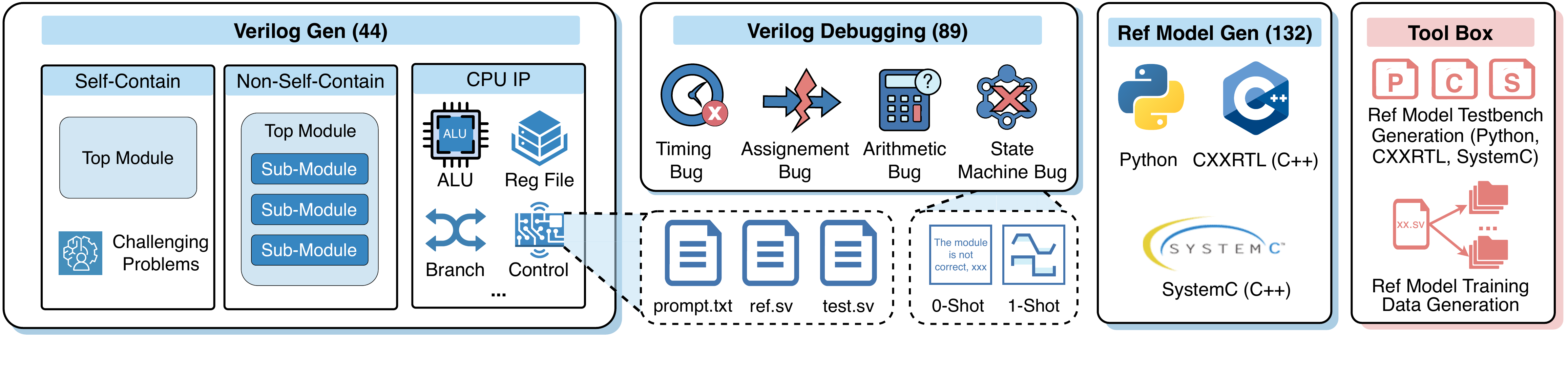}
    \caption{Overview of our ~\xname}
    \label{fig:overview}
\end{figure*}

\textbf{1. Overly simplified benchmarks overestimate LLMs' capabilities for industrial deployment.}
Existing Verilog benchmarks contain only simple functional modules ranging from 10 to 76 lines, with all modules being self-contained and never instantiating sub-modules.
This contrasts sharply with industrial reality, where Verilog modules typically exceed 10,000 lines with complex functionality and employ hierarchical designs that instantiate and interact with multiple sub-modules for code reuse and systematic verification~\cite{keutzer2000system, jin2025realbench}.
The gap is further widened by the source of these modules: most samples in existing benchmarks are derived from coding contests rather than real industrial projects, resulting in significant discrepancies in both functionality and coding style compared to production code.
Consequently, even if LLMs excel in existing benchmarks, they may still perform poorly in industrial-level deployment due to these substantial gaps between benchmark scenarios and production requirements.


\textbf{2. The exclusive focus on Verilog code generation overlooks LLMs’ debugging capabilities, which are even more critical for industrial development.}
The semiconductor industry's extreme caution stems from the fact that design bugs can result in billions of dollars in losses~\cite{price1995pentium}, making companies reluctant to adopt LLMs for complete code generation due to their inability to produce consistently bug-free code. 
Given these constraints, LLM-assisted debugging represents a more practical and immediately deployable alternative that integrates seamlessly into existing workflows without disrupting established safety protocols. 
Despite debugging being more urgent and realistic than full generation, existing benchmarks focus exclusively on code generation, creating a critical gap in evaluating LLMs' practical utility for industrial integration.

\textbf{3. The narrow focus on Verilog generation neglects reference model evaluation, which is equally vital for industrial chip design.}
Reference models, typically written in high-level languages (e.g., SystemC~\cite{panda2001systemc}, Python, and CXXRTL~\cite{wolf2013yosys}), serve as predictors of hardware behavior and enable cross-comparison to eliminate potential bugs~\cite{jiang2020pymtl3, zhao2025pro, hung2004reference}. 
In industrial workflows, reference model generation is even more resource-intensive than Verilog design, reflected in a 1:1 - 5:1 ratio of verification engineers (write reference model) to design engineers (write Verilog)~\cite{foster20222022}.
Despite this importance, existing benchmarks completely ignore this task, leading researchers to overlook this critical area and limiting LLM's potential to enhance the chip design industry.

To address these gaps, we propose \xname, a comprehensive benchmark designed to 
bridge the gap between academic datasets and industrial complexity. 
By expanding evaluation beyond simple generation to include debugging and reference modeling, \xname~provides a rigorous assessment of LLM capabilities. 
We summarize our contributions as follows:
\begin{itemize}
\item \textbf{A Challenging and Realistic Verilog Generation Benchmark.} 
    We compile 44 modules from open-source CPU IPs and competitive platforms, categorized into self-contained, non-self-contained, and real-world CPU submodules.
    As shown in \autoref{tab:line_cell_compare}, our dataset features 3.8$\times$ longer code length and 13.9$\times$ more cells than VerilogEval.
    Test results show that the most powerful multi-agent system MAGE achieves only 37\% pass rate, significantly lower than its 95\% pass rate on VerilogEval, demonstrating our benchmark's potential as a robust milestone for future research.

\item \textbf{A Systematic Verilog Debugging Benchmark.} 
We introduce a debugging suite with 89 test cases covering four error types: timing, arithmetic, assignment, and state machine bugs. 
By manually injecting these errors, we create a controlled environment to evaluate LLM repair capabilities, 
Test results show that LLMs generally demonstrate 5\%-20\% higher pass rates on debugging than code generation, demonstrating their potential to be integrated into industrial workflows.

    \item \textbf{A Heterogeneous Reference Model Generation Benchmark.} 
    We introduce 132 samples (44×3), enabling evaluation of reference model generation across Python, SystemC, and CXXRTL.
    Our self-developed framework supports systematic comparison between reference models and golden Verilog implementations, and is extensible to any existing Verilog codebase.
    Evaluation reveals significant capability gaps, with top-tier Claude-4.5-opus~\cite{anthropic2025opus} achieving only 15.93\% accuracy in Python, demonstrating substantial performance decline compared to other Python tasks and highlighting LLMs' unexplored potential in this critical area.
    
    \item \textbf{Automated Toolbox for Reference Model Verification and Training Dataset Generation.} 
    To facilitate future research on reference model generation, we release an automated toolbox alongside \xname. This tool automates the verification of LLM-generated reference models and facilitates the generation of high-quality training datasets from existing Verilog datasets.
    We tested our toolbox on a subset of QiMeng CodeV-R1 training dataset~\cite{zhu2025qimeng} and have successfully generated 2206 high-quality Python reference models from 10,000 Verilog training data.
\end{itemize}

\section{~\xname}
In this section, we will introduce the components and workflow of ~\xname{} in detail.

\subsection{Overview}
Our proposed \xname{} extends beyond a simple Verilog coding testbench to comprehensively evaluate LLMs across multiple aspects of the chip design workflow. Through this benchmark, we emphasize the importance of debugging and reference model generation, capabilities that could elevate AI-aided chip design to new heights.
As illustrated in ~\autoref{fig:overview} and ~\autoref{tab:line_cell_compare}, \xname{} encompasses three core tasks spanning over 264 test cases: 44 structurally complex and challenging Verilog generation cases, 89 Verilog debugging cases, and 132 reference model generation cases distributed across Python, SystemC, and CXXRTL (extension to infinite cases with our proposed toolbox). We summarized details of our modules in ~\autoref{tab:module_descriptions_part1} in the Appendix.

In addition to the comprehensive test suite, we provide an integrated toolbox for reference model verification and automated training dataset generation. Given the current shortage of high-quality reference model training data, this toolbox serves as a valuable resource for the research community. To validate its effectiveness, we deployed the toolbox with DeepSeek V3.2 and successfully generated over 2,200 high-quality Python reference models from 10,000 Verilog training cases~\cite{zhu2025qimeng}. Given the abundance of existing Verilog training datasets, including QiMeng CodeV (87k samples)\cite{zhu2025qimeng}, Pyranet (692k samples)\cite{nadimi2025pyranet}, and VeriGen (50k samples)~\cite{thakur2024verigen}, our toolbox has the potential to generate extensive reference model datasets at scale, thereby significantly benefiting the research community.

\begin{table}[t]
    \centering
    \caption{Comparison with previous benchmarks.}
    \label{tab:line_cell_compare}
    
    \definecolor{mygreen}{HTML}{34A853}
    \definecolor{myred}{HTML}{EA4335}
    \definecolor{rowgray}{gray}{0.92}
    \definecolor{rowblue}{HTML}{E8F0FE}
    
    \resizebox{\columnwidth}{!}{%
        \begin{tabular}{l cccc cc} 
            \toprule
            \multirow{2.5}{*}{\textbf{Benchmarks}} & \multicolumn{4}{c}{\textbf{Verilog Gen}} & \textbf{V-Debug} & \textbf{Ref-Gen} \\ 
            \cmidrule(lr){2-5} \cmidrule(lr){6-6} \cmidrule(lr){7-7}
            & Code & Cells & Subs & Tests & Tests & Tests \\ 
            \midrule
            
            Thakur et al. & 16.6 & 7.9 & 0 & 17 & \textcolor{myred}{\ding{55}} & \textcolor{myred}{\ding{55}} \\
            RTLLMV2 & 46.1 & 33.7 & 0.2 & 50 & \textcolor{myred}{\ding{55}} & \textcolor{myred}{\ding{55}} \\
            VerilogEvalV2 & 16.1 & 31.4 & 0 & 156 & \textcolor{myred}{\ding{55}} & \textcolor{myred}{\ding{55}} \\
            
            \midrule
            \rowcolor{rowgray}
            Self-Contn (ours) & 47.8 & 323.3 & 0 & 29 & \textcolor{mygreen}{\ding{51}} & \textcolor{mygreen}{\ding{51}} \\
            \rowcolor{rowgray}
            Not-Self-Contn (ours) & 119.5 & 361.2 & 1.8 & 6 & \textcolor{myred}{\ding{55}} & \textcolor{mygreen}{\ding{51}} \\
            \rowcolor{rowgray}
            Cpu-Design (ours) & 68.1 & 862.3 & 0 & 9 & \textcolor{mygreen}{\ding{51}} & \textcolor{mygreen}{\ding{51}} \\
            
            \midrule
            \rowcolor{rowblue}
            \textbf{ChipBench (ours)} & \textbf{61.7} & \textbf{438.7} & \textbf{0.25} & \textbf{44} & \textbf{89} & \textbf{132+} \\ 
            
            \arrayrulecolor{black} 
            \midrule 
            
            \rowcolor{rowblue}
            \textit{Total Test Cases} & \multicolumn{6}{c}{\textbf{264+}} \\
            \bottomrule
        \end{tabular}%
    }
    
    \vspace{3pt} 
    \raggedright 
    \footnotesize 
\end{table}

\subsection{Verilog Generation Tasks}
We developed a dataset of high-quality golden Verilog code sourced from open-source Online Judge (OJ) platforms and established CPU IP projects. 
Each test case was carefully selected to ensure appropriate difficulty levels that are neither trivial nor overly challenging. For each test case, we manually wrote description prompts rather than using LLM-generated ones to ensure accurate functional descriptions, consistent with recommendations from VerilogEval V2~\cite{pinckney2025revisiting}.

To conduct a comprehensive evaluation of LLM coding capabilities, we categorize the test cases into three distinct categories. The first category consists of self-contained module tests, where modules operate independently without instantiating external sub-modules. While this format aligns with existing benchmarks like VerilogEval, as shown in ~\autoref{tab:line_cell_compare}, \xname{} significantly elevates the complexity, with samples averaging 47.8 lines of code and 323.3 cells compared to approximately 16.1 lines and 31.4 cells in VerilogEval V2, presenting a more rigorous challenge to the model's logic synthesis capabilities.

The second category covers hierarchical modules that are non-self-contained. These cases require the top-level module to instantiate and interact with sub-modules, reflecting real-world industrial scenarios where coding standards prioritize modularity and reuse to minimize verification overhead. To assist LLMs in understanding interface requirements, we include both the description and source code of sub-modules within the prompt. We only require LLMs to complete the top module, since all models failed when tasked with generating complete hierarchical designs, making such tests prohibitively challenging.

The third category involves CPU IP modules sourced directly from open-source CPU projects to reflect the demands of high-performance computing design. 
We manually selected representative functional units, including ALUs, controllers, branch predicters, etc, with moderate complexity from larger project contexts.
These test cases assess the LLM's ability to generate functional Verilog code that meets professional engineering standards and real-world deployment requirements.

For every test case in \xname{}, we provide a golden Verilog implementation, a manually developed test file, and a high-quality prompt written by domain experts. The test files utilize a robust combination of directed tests and constrained random testing (over 1,000 iterations) to ensure high-fidelity evaluation.

\definecolor{headergray}{RGB}{90, 90, 90}
\definecolor{hlblue}{RGB}{173, 216, 230}
\definecolor{hlred}{RGB}{255, 182, 182}
\definecolor{hlgreen}{RGB}{182, 255, 182}
\definecolor{hlyellow}{RGB}{255, 255, 0}

\newlength{\boxwidth}
\setlength{\boxwidth}{6.5cm}
\newlength{\innerwidth}
\setlength{\innerwidth}{6.1cm}  

\tcbset{
    codebox/.style={
        colframe=gray!50,
        colback=white,
        boxrule=0.5pt,
        arc=0pt,
        left=2pt, right=2pt, top=2pt, bottom=2pt,
        fontupper=\ttfamily\fontsize{10}{12}\selectfont,
        width=\boxwidth,
        height=14cm,
        valign=top,
    },
    headerbox/.style={
        colback=headergray,
        colframe=headergray,
        colupper=white,
        fontupper=\bfseries\large,
        halign=center,
        arc=0pt,
        boxrule=0pt,
        width=\boxwidth,
    }
}

\newcommand{\hlb}[1]{\colorbox{hlblue}{\makebox[\innerwidth][l]{#1}}}   
\newcommand{\hlr}[1]{\colorbox{hlred}{\makebox[\innerwidth][l]{#1}}}    
\newcommand{\hlg}[1]{\colorbox{hlgreen}{\makebox[\innerwidth][l]{#1}}}  
\newcommand{\hly}[1]{\colorbox{hlyellow}{\makebox[\innerwidth][l]{#1}}}
\newcommand{\kw}[1]{\textcolor{blue}{#1}}     
\renewcommand{\ln}[1]{\textcolor{gray}{#1}}     

\begin{figure*}[htbp]
\centering
\resizebox{\textwidth}{!}{%
    \begin{tabular}{*{5}{c}}
\begin{minipage}{\boxwidth}
\begin{tcolorbox}[headerbox]
Original
\end{tcolorbox}
\begin{tcolorbox}[codebox]
\ln{1} \kw{module} TopModule(\\
\ln{2} \kw{input} clk,\\
\ln{3} \kw{input} rst\_n, \\
\ln{4} \kw{output reg} red, \\
\ln{......}\\
\ln{10} );\\
\ln{......}\\
\ln{38} s1\_red: \\
\ln{39} \kw{begin}: \\
\ln{40} p\_red <= 1'b1;\\
\hly{\ln{41} p\_green <= 1'b0;}\\
\ln{42} p\_yellow <= 1'b0;\\
\ln{43} \kw{if} (cnt == 3)\\
\hlb{\ln{44} state <= s2\_green;}\\
\ln{45} \kw{else}\\
\ln{46} state <= s1\_red;\\
\ln{47} \kw{end} \\
\ln{......}\\
\ln{71} \kw{always} @(\kw{posedge} clk \kw{or} \kw{negedge} rst\_n)\\
\ln{72} \kw{if}(!rst\_n)\\
\ln{73} cnt <= 7'd10;\\
\hlg{\ln{76} \kw{else if} (!green\&\&p\_green)}\\
\ln{77} cnt <= 7'd60;\\
\ln{......} \\
\ln{85} \kw{always} @(\kw{posedge} clk \kw{or} \kw{negedge} rst\_n) \\
\ln{86} \kw{if}(!rst\_n)\\
\ln{87} \kw{begin}\\
\hlr{\ln{88} yellow <= 1'd0;}\\
\ln{......} \\
\ln{99} \kw{endmodule}
\end{tcolorbox}
\end{minipage}

&


\begin{minipage}{\boxwidth}
\begin{tcolorbox}[headerbox]
Arithmetic Bug (24)
\end{tcolorbox}
\begin{tcolorbox}[codebox]
\ln{1} \kw{module} TopModule(\\
\ln{2} \kw{input} clk,\\
\ln{3} \kw{input} rst\_n, \\
\ln{4} \kw{output reg} red, \\
\ln{......}\\
\ln{10} );\\
\ln{......}\\
\ln{38} s1\_red: \\
\ln{39} \kw{begin}: \\
\ln{40} p\_red <= 1'b1;\\
\ln{41} p\_green <= 1'b1;\\
\ln{42} p\_yellow <= 1'b0;\\
\ln{43} \kw{if} (cnt == 3)\\
\ln{44} state <= s2\_green;\\
\ln{45} \kw{else}\\
\ln{46} state <= s1\_red;\\
\ln{47} \kw{end} \\
\ln{......}\\
\ln{71} \kw{always} @(\kw{posedge} clk \kw{or} \kw{negedge} rst\_n)\\
\ln{72} \kw{if}(!rst\_n)\\
\ln{73} cnt <= 7'd10;\\
\hlg{\ln{76} \kw{else if} (green\&\&p\_green)}\\
\ln{77} cnt <= 7'd60;\\
\ln{......} \\
\ln{85} \kw{always} @(\kw{posedge} clk \kw{or} \kw{negedge} rst\_n) \\
\ln{86} \kw{if}(!rst\_n)\\
\ln{87} \kw{begin}\\
\ln{88} yellow <= 1'd0;\\
\ln{......} \\
\ln{99} \kw{endmodule}
\end{tcolorbox}
\end{minipage}

&

\begin{minipage}{\boxwidth}
\begin{tcolorbox}[headerbox]
Assignment Bug (30)
\end{tcolorbox}
\begin{tcolorbox}[codebox]
\ln{1} \kw{module} TopModule(\\
\ln{2} \kw{input} clk,\\
\ln{3} \kw{input} rst\_n, \\
\ln{4} \kw{output reg} red, \\
\ln{......}\\
\ln{10} );\\
\ln{......}\\
\ln{38} s1\_red: \\
\ln{39} \kw{begin}: \\
\ln{40} p\_red <= 1'b1;\\
\hly{\ln{41} p\_green <= 1'b1;}\\
\ln{42} p\_yellow <= 1'b0;\\
\ln{43} \kw{if} (cnt == 3)\\
\ln{44} state <= s2\_green;\\
\ln{45} \kw{else}\\
\ln{46} state <= s1\_red;\\
\ln{47} \kw{end} \\
\ln{......}\\
\ln{71} \kw{always} @(\kw{posedge} clk \kw{or} \kw{negedge} rst\_n)\\
\ln{72} \kw{if}(!rst\_n)\\
\ln{73} cnt <= 7'd10;\\
\ln{76} \kw{else if} (!green\&\&p\_green)\\
\ln{77} cnt <= 7'd60;\\
\ln{......} \\
\ln{85} \kw{always} @(\kw{posedge} clk \kw{or} \kw{negedge} rst\_n) \\
\ln{86} \kw{if}(!rst\_n)\\
\ln{87} \kw{begin}\\
\ln{88} yellow <= 1'd0;\\
\ln{......} \\
\ln{99} \kw{endmodule}
\end{tcolorbox}
\end{minipage}

&

\begin{minipage}{\boxwidth}
\begin{tcolorbox}[headerbox]
Timing Bug (29)
\end{tcolorbox}
\begin{tcolorbox}[codebox]
\ln{1} \kw{module} TopModule(\\
\ln{2} \kw{input} clk,\\
\ln{3} \kw{input} rst\_n, \\
\ln{4} \kw{output reg} red, \\
\ln{......}\\
\ln{10} );\\
\ln{......}\\
\ln{38} s1\_red: \\
\ln{39} \kw{begin}: \\
\ln{40} p\_red <= 1'b1;\\
\ln{41} p\_green <= 1'b0;\\
\ln{42} p\_yellow <= 1'b0;\\
\ln{43} \kw{if} (cnt == 3)\\
\ln{44} state <= s2\_green;\\
\ln{45} \kw{else}\\
\ln{46} state <= s1\_red;\\
\ln{47} \kw{end} \\
\ln{......}\\
\ln{71} \kw{always} @(\kw{posedge} clk \kw{or} \kw{negedge} rst\_n)\\
\ln{72} \kw{if}(!rst\_n)\\
\ln{73} cnt <= 7'd10;\\
\ln{76} \kw{else if} (!green\&\&p\_green)\\
\ln{77} cnt <= 7'd60;\\
\ln{......} \\
\ln{85} \kw{always} @(*) \\
\ln{86} \kw{if}(!rst\_n)\\
\ln{87} \kw{begin}\\
\hlr{\ln{88} yellow = 1'd0;}\\
\ln{......} \\
\ln{99} \kw{endmodule}
\end{tcolorbox}
\end{minipage}

&

\begin{minipage}{\boxwidth}
\begin{tcolorbox}[headerbox]
State Machine Bug (6)
\end{tcolorbox}
\begin{tcolorbox}[codebox]
\ln{1} \kw{module} TopModule(\\
\ln{2} \kw{input} clk,\\
\ln{3} \kw{input} rst\_n, \\
\ln{4} \kw{output reg} red, \\
\ln{......}\\
\ln{10} );\\
\ln{......}\\
\ln{38} s1\_red: \\
\ln{39} \kw{begin}: \\
\ln{40} p\_red <= 1'b1;\\
\ln{41} p\_green <= 1'b0;\\
\ln{42} p\_yellow <= 1'b0;\\
\ln{43} \kw{if} (cnt == 3)\\
\hlb{\ln{44} state <= s3\_yellow;}\\
\ln{45} \kw{else}\\
\ln{46} state <= s1\_red;\\
\ln{47} \kw{end} \\
\ln{......}\\
\ln{71} \kw{always} @(\kw{posedge} clk \kw{or} \kw{negedge} rst\_n)\\
\ln{72} \kw{if}(!rst\_n)\\
\ln{73} cnt <= 7'd10;\\
\ln{76} \kw{else if} (!green\&\&p\_green)\\
\ln{77} cnt <= 7'd60;\\
\ln{......} \\
\ln{85} \kw{always} @(\kw{posedge} clk \kw{or} \kw{negedge} rst\_n) \\
\ln{86} \kw{if}(!rst\_n)\\
\ln{87} \kw{begin}\\
\ln{88} yellow <= 1'd0;\\
\ln{......} \\
\ln{99} \kw{endmodule}
\end{tcolorbox}
\end{minipage}

\end{tabular}
}

\caption{Demonstration of four types of bugs injected into original golden modules.}
\label{fig: bugs_demo}
\end{figure*}

\subsection{Verilog Debugging Tasks}
To evaluate LLMs' error correction capabilities, we constructed debugging tasks by manually injecting faults into golden Verilog modules. We identified four categories of bugs most prevalent in hardware engineering: arithmetic bugs, assignment bugs, timing bugs, and state machine bugs.
As shown in ~\autoref{fig: bugs_demo}, arithmetic bugs involve errors in operator usage, such as substituting multiplication operators for addition operators. Assignment bugs refer to incorrect constant value assignments to wires or registers. Timing bugs occur specifically in sequential logic, manifesting as values assigned in incorrect clock cycles or misuse of blocking versus non-blocking assignments. State machine bugs represent logic errors within state transition mechanisms, where the transition logic itself is flawed.

To simulate realistic debugging workflows, we designed two distinct testing modes: zero-shot and one-shot evaluation. The zero-shot test provides the model with the module description and buggy implementation, indicating that an error exists without providing localization details. The one-shot test provides identical information but supplements it with simulation waveform data (.vcd files). This replicates industrial workflows where design and verification engineers analyze simulation waveforms to isolate and diagnose faults.


\begin{figure}[t]
    \centering
    \includegraphics[width=0.99\columnwidth]{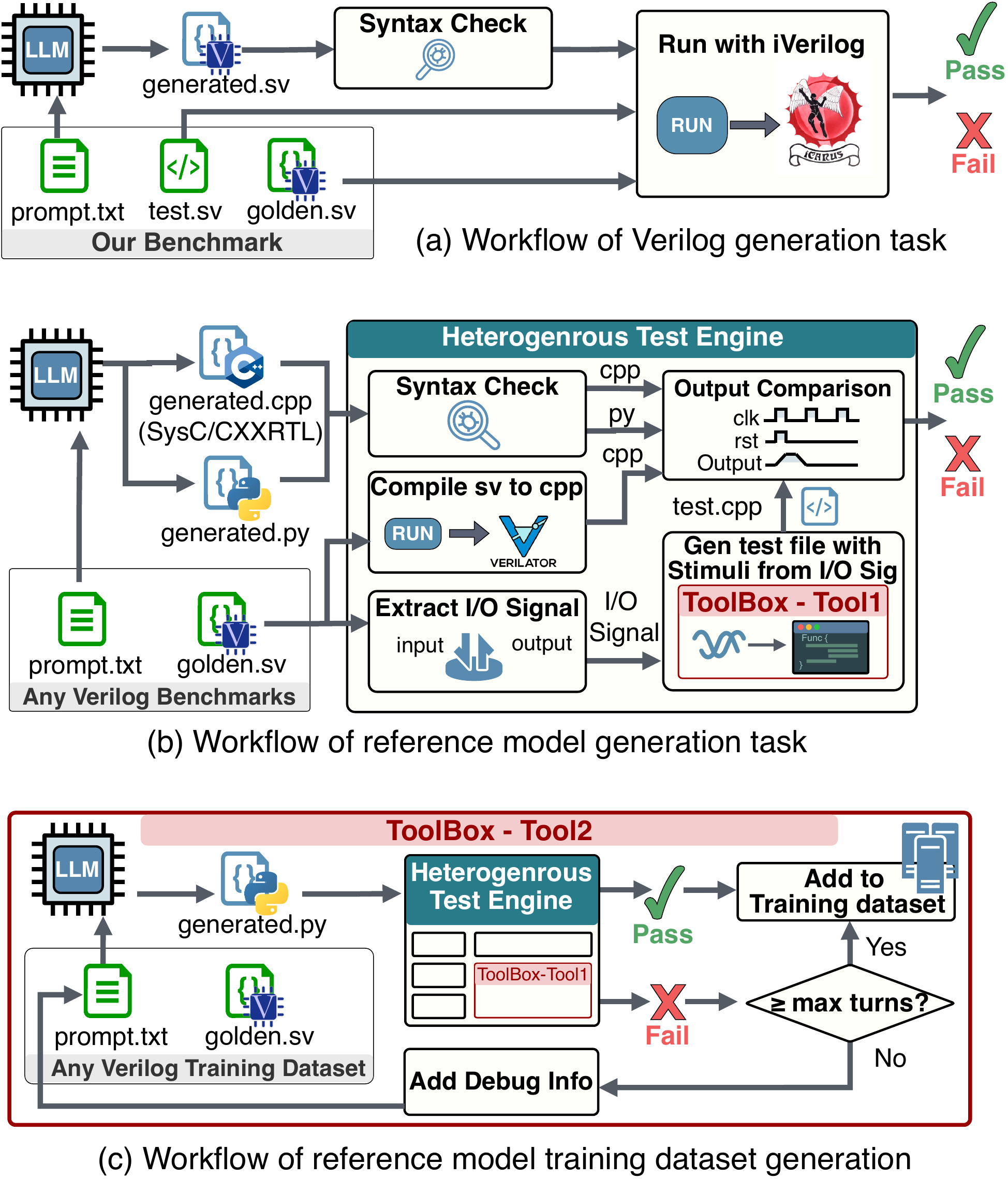}
    \caption{Workflow of ~\xname{} across different tasks: (a) evaluating LLM-generated Verilog code, (b) assessing LLM-generated reference models (supporting Python, SystemC, and CXXRTL), and (c) utilizing the toolbox to generate high quality training datasets for reference models (supporting Python, SystemC, and CXXRTL, with Python shown as an example)}
    \label{fig:workflow}
\end{figure}

\begin{figure}[t]
    \centering
    \includegraphics[width=0.9\columnwidth]{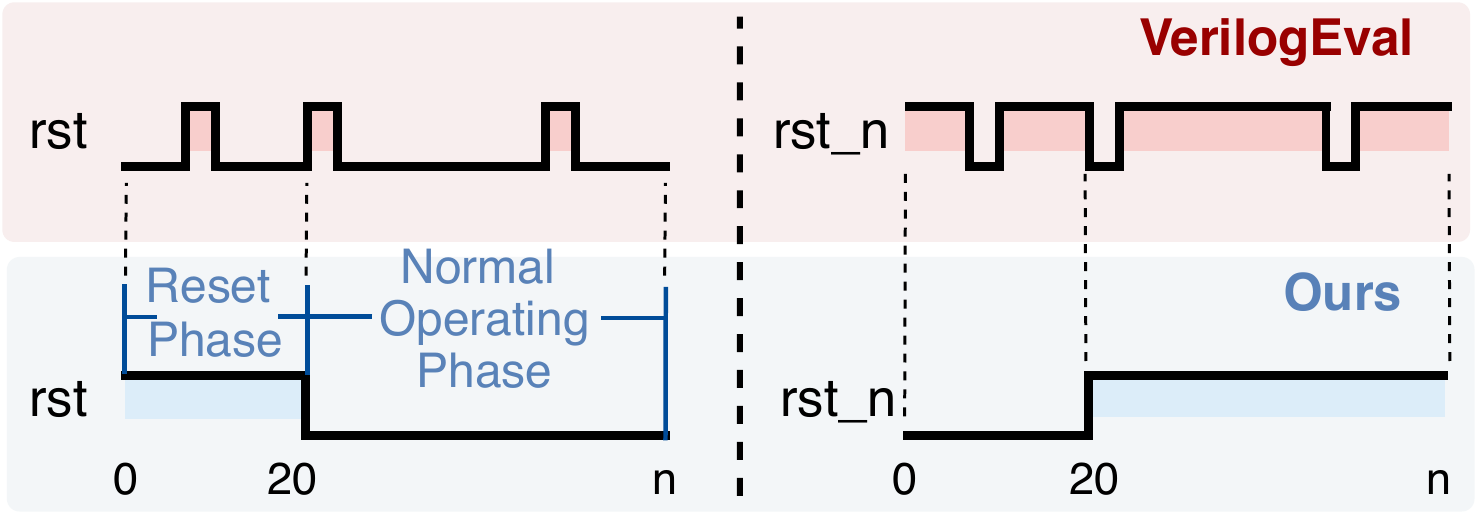}
    \caption{Comparison of reset signal patterns between our approach and VerilogEval. We toggle reset once and only compare outputs during the normal operating phase to better reflect real-world verification standards and eliminate false negatives.}
    \label{fig:rst}
\end{figure}

\subsection{Framework workflow}
We demonstrate the workflow for the Verilog generation task and the reference model generation task in ~\autoref{fig:workflow}(a)(b).

\textbf{Verilog Generation.} Our evaluation framework is built upon VerilogEval, ensuring full functional compatibility. As shown in \autoref{fig:workflow}(a), each test case comprises three files: a \texttt{prompt.txt} file describing the task, a \texttt{test.sv} harness for evaluation, and a \texttt{golden.sv} serving as the ground truth. Upon generating a solution (\texttt{generated.sv}), the system first performs a syntax check to ensure compilability. Subsequently, we execute \texttt{test.sv} using iVerilog to compare the outputs of the generated code against the golden reference. To ensure reliability, the test harness employs a combination of manually designed corner cases and over 1,000 randomly generated stimuli.

\textbf{Reference Model Generation.} Our framework supports verification of LLM-generated reference models through a custom Heterogeneous Test Engine (HTE).  
As illustrated in \autoref{fig:workflow}(b), this workflow requires a module description file \texttt{prompt.txt} and a golden implementation file \texttt{golden.sv} as input and supports multiple reference model languages, including Python and C++ (via SystemC or CXXRTL extensions).
Once the LLM generates a reference model, HTE performs a syntax check followed by functional verification. 
To validate the reference model, we extract I/O signals from \texttt{golden.sv} and generate a heterogeneous test harness (\texttt{test.cpp}). 
Simultaneously, the \texttt{golden.sv} is compiled into C++ using Verilator~\cite{verilator}. 
The test harness then compares the outputs of the LLM-generated reference model against the transferred golden model using over 1,000 random stimuli. This automated flow is theoretically extensible to any testbench, as long as a prompt and a golden Verilog implementation are provided.

\subsection{Toolbox}
\xname{} includes a specialized Toolbox to facilitate future research in reference model generation, comprising two primary tools shown in \autoref{fig:workflow}(b) and (c).

\textbf{Tool 1: Automatic Testbench Generation.} This tool automates the creation of test files of LLM-generated references based on the I/O signals extracted from the golden Verilog. It injects random input stimuli to robustly compare the golden implementation with the generated reference model. 

For input signals, we carefully designed reset signal patterns to align with industrial verification standards.
As shown in \autoref{fig:rst}, unlike VerilogEval test cases that toggle reset signals frequently, our tool identifies the reset polarity (active-high \texttt{rst} or active-low \texttt{rst\_n}) and asserts system reset for the initial 20 cycles before transitioning to continuous operation.
To prevent spurious failures, we mask output comparisons during the reset phase and only compare outputs during normal operation. This design addresses the fundamental discrepancy between Verilog, which may output `X' (unknown) states before initialization, and high-level languages like Python/C++, which lack equivalent unknown states. This approach eliminates false negatives and better reflects real-world verification practices.

\textbf{Tool 2: Automatic Training Dataset Generation.} To address the shortage of training data for reference model generation, we extended the HTE to create an automated training dataset generation pipeline. 
As shown in \autoref{fig:workflow}(c), this tool is compatible with any Verilog training dataset containing prompts (\texttt{prompt.txt}) and golden implementations (\texttt{golden.sv}).

Our pipeline employs a multi-turn feedback loop: if an LLM-generated reference model fails verification, the error information is fed back to the model for self-correction. We validated this workflow using DeepSeek V3.2 on 10,000 randomly selected samples from the CodeV-R1 dataset. The process successfully generated 2,206 verified, high-quality reference models, demonstrating the tool's effectiveness in constructing robust training data for AI-aided chip design.

\section{Evaluation}




\subsection{Evaluation Setup}
\textbf{Models.} We evaluate \xname~across multiples state-of-the-art LLMs, including GPT (3.5, 4o, 5, 5.2), 
Claude (Haiku, Opus, Sonnet), Gemini(2.5-pro, 3-flash), Llama3.1 (8B, 70B), 
DeepSeek-V3.2(coder, reasoner). The detailed version information for each model is listed in ~\autoref{tab:model_details} in the Appendix.
For all models, we use the 
hyperparameters recommended by VerilogEval: \texttt{temperature=0.85} and 
\texttt{top\_p=0.95}.

\textbf{Multi-Agent Workflow.} In addition to single-model evaluations, we test 
MAGE with DeepSeek-V3.2 backbone, a state-of-the-art multi-agent framework for Verilog generation. MAGE 
employs dedicated agents for sampling, debugging, and decision-making to enhance 
code quality.

\textbf{Metrics.} We report pass@1, pass@5, and pass@10 for four distinct tasks: 
Verilog generation, Verilog debugging, and Python reference model generation. 
For the multi-agent workflow, we report only 
pass@1, as the workflow inherently involves multi-turn iterations. Additionally, 
we track token consumption to estimate the API costs associated with running 
our benchmark.

\begin{table*}[t]
\centering
\renewcommand{\arraystretch}{1.1}
\caption{Test results for Verilog generation (Red color denotes low pass rate, which green color denotes high)}
\label{tab:verilog_gen}

\resizebox{0.83\textwidth}{!}{%
\begin{tabular}{lcccccccccccc}
\toprule
\multirow{2}{*}{\textbf{Model/Dataset}} & \multicolumn{3}{c}{\textbf{CPU IP}} & \multicolumn{3}{c}{\textbf{Non-Self-Contained}} & \multicolumn{3}{c}{\textbf{Self-Contained}} & \multicolumn{3}{c}{\textbf{Average}} \\
\cmidrule(lr){2-4} \cmidrule(lr){5-7} \cmidrule(lr){8-10} \cmidrule(lr){11-13}
 & \textbf{pass@1} & \textbf{pass@5} & \textbf{pass@10} & \textbf{pass@1} & \textbf{pass@5} & \textbf{pass@10} & \textbf{pass@1} & \textbf{pass@5} & \textbf{pass@10} & \textbf{pass@1} & \textbf{pass@5} & \textbf{pass@10} \\
\midrule

\textbf{Claude 4.5 Haiku} & 
\cellcolor[HTML]{DE8C89}11.11 & 
\cellcolor[HTML]{DE8C89}11.11 & 
\cellcolor[HTML]{DE8C89}11.11 & 
\cellcolor[HTML]{E19791}16.67 & 
\cellcolor[HTML]{EBB3AA}33.33 & 
\cellcolor[HTML]{EBB3AA}33.33 & 
\cellcolor[HTML]{E9ADA5}30 & 
\cellcolor[HTML]{EEBCB2}36.67 & 
\cellcolor[HTML]{EEBCB2}36.67 & 
\cellcolor[HTML]{E29B95}19.26 & 
\cellcolor[HTML]{E7A8A1}27.04 & 
\cellcolor[HTML]{E7A8A1}27.04 \\

\textbf{Claude 4.5 Opus} & 
\cellcolor[HTML]{E4A09A}22.22 & 
\cellcolor[HTML]{E4A09A}22.22 & 
\cellcolor[HTML]{E4A09A}22.22 & 
\cellcolor[HTML]{EBB3AA}33.33 & 
\cellcolor[HTML]{FAE1D2}50 & 
\cellcolor[HTML]{FAE1D2}50 & 
\cellcolor[HTML]{EEBCB2}36.67 & 
\cellcolor[HTML]{F4CEC2}43.33 & 
\cellcolor[HTML]{FAE1D2}50 & 
\cellcolor[HTML]{E9AFA6}30.74 & 
\cellcolor[HTML]{EFC1B6}38.52 & 
\cellcolor[HTML]{F1C7BC}40.74 \\

\textbf{Claude 4.5 Sonnet} & 
\cellcolor[HTML]{DE8C89}11.11 & 
\cellcolor[HTML]{DE8C89}11.11 & 
\cellcolor[HTML]{DE8C89}11.11 & 
\cellcolor[HTML]{E19791}16.67 & 
\cellcolor[HTML]{EBB3AA}33.33 & 
\cellcolor[HTML]{EBB3AA}33.33 & 
\cellcolor[HTML]{DE8A87}10 & 
\cellcolor[HTML]{E39C96}20 & 
\cellcolor[HTML]{E7A8A0}26.67 & 
\cellcolor[HTML]{DF8F8B}12.59 & 
\cellcolor[HTML]{E49F99}21.48 & 
\cellcolor[HTML]{E5A39C}23.70 \\

\textbf{DeepSeek-Coder} & 
\cellcolor[HTML]{DE8C89}11.11 & 
\cellcolor[HTML]{DE8C89}11.11 & 
\cellcolor[HTML]{DE8C89}11.11 & 
\cellcolor[HTML]{FAE1D2}50 & 
\cellcolor[HTML]{FAE1D2}50 & 
\cellcolor[HTML]{FAE1D2}50 & 
\cellcolor[HTML]{E7A8A0}26.67 & 
\cellcolor[HTML]{EEBCB2}36.67 & 
\cellcolor[HTML]{F4CEC2}43.33 & 
\cellcolor[HTML]{E8ACA4}29.26 & 
\cellcolor[HTML]{EAB2A9}32.59 & 
\cellcolor[HTML]{ECB7AE}34.81 \\

\textbf{DeepSeek-R1} & 
\cellcolor[HTML]{DE8C89}11.11 & 
\cellcolor[HTML]{DE8C89}11.11 & 
\cellcolor[HTML]{DE8C89}11.11 & 
\cellcolor[HTML]{EBB3AA}33.33 & 
\cellcolor[HTML]{FAE1D2}50 & 
\cellcolor[HTML]{FAE1D2}50 & 
\cellcolor[HTML]{E7A8A0}26.67 & 
\cellcolor[HTML]{F1C5BA}40 & 
\cellcolor[HTML]{EFE2CF}53.33 & 
\cellcolor[HTML]{E5A39C}23.70 & 
\cellcolor[HTML]{EBB4AB}33.70 & 
\cellcolor[HTML]{EFC0B6}38.15 \\

\textbf{Gemini 2.5 Pro} & 
\cellcolor[HTML]{DA7777}0 & 
\cellcolor[HTML]{DE8C89}11.11 & 
\cellcolor[HTML]{DE8C89}11.11 & 
\cellcolor[HTML]{DA7777}0 & 
\cellcolor[HTML]{E19791}16.67 & 
\cellcolor[HTML]{E19791}16.67 & 
\cellcolor[HTML]{DB7D7C}3.33 & 
\cellcolor[HTML]{DE8A87}10 & 
\cellcolor[HTML]{DF908C}13.33 & 
\cellcolor[HTML]{DA7978}1.11 & 
\cellcolor[HTML]{DF8F8B}12.59 & 
\cellcolor[HTML]{DF918D}13.70 \\

\textbf{Gemini 3 Flash} & 
\cellcolor[HTML]{E4A09A}22.22 & 
\cellcolor[HTML]{E4A09A}22.22 & 
\cellcolor[HTML]{E4A09A}22.22 & 
\cellcolor[HTML]{EBB3AA}33.33 & 
\cellcolor[HTML]{FAE1D2}50 & 
\cellcolor[HTML]{FAE1D2}50 & 
\cellcolor[HTML]{EEBCB2}36.67 & 
\cellcolor[HTML]{F7D7CA}46.67 & 
\cellcolor[HTML]{F7D7CA}46.67 & 
\cellcolor[HTML]{E9AFA6}30.74 & 
\cellcolor[HTML]{F0C4B9}39.63 & 
\cellcolor[HTML]{F0C4B9}39.63 \\

\textbf{GPT-3.5 Turbo} & 
\cellcolor[HTML]{DE8C89}11.11 & 
\cellcolor[HTML]{DE8C89}11.11 & 
\cellcolor[HTML]{DE8C89}11.11 & 
\cellcolor[HTML]{DA7777}0 & 
\cellcolor[HTML]{DA7777}0 & 
\cellcolor[HTML]{DA7777}0 & 
\cellcolor[HTML]{DF908C}13.33 & 
\cellcolor[HTML]{E7A8A0}26.67 & 
\cellcolor[HTML]{E7A8A0}26.67 & 
\cellcolor[HTML]{DD8684}8.15 & 
\cellcolor[HTML]{DF8F8B}12.59 & 
\cellcolor[HTML]{DF8F8B}12.59 \\

\textbf{GPT-4o} & 
\cellcolor[HTML]{DE8C89}11.11 & 
\cellcolor[HTML]{DE8C89}11.11 & 
\cellcolor[HTML]{DE8C89}11.11 & 
\cellcolor[HTML]{DA7777}0 & 
\cellcolor[HTML]{DA7777}0 & 
\cellcolor[HTML]{DA7777}0 & 
\cellcolor[HTML]{E39C96}20 & 
\cellcolor[HTML]{EBB3AA}33.33 & 
\cellcolor[HTML]{F7D7CA}46.67 & 
\cellcolor[HTML]{DE8B87}10.37 & 
\cellcolor[HTML]{E0938F}14.81 & 
\cellcolor[HTML]{E29B95}19.26 \\

\textbf{GPT-5} & 
\cellcolor[HTML]{DE8C89}11.11 & 
\cellcolor[HTML]{DE8C89}11.11 & 
\cellcolor[HTML]{E4A09A}22.22 & 
\cellcolor[HTML]{EBB3AA}33.33 & 
\cellcolor[HTML]{EBB3AA}33.33 & 
\cellcolor[HTML]{EBB3AA}33.33 & 
\cellcolor[HTML]{E9ADA5}30 & 
\cellcolor[HTML]{FAE1D2}50 & 
\cellcolor[HTML]{EFE2CF}53.33 & 
\cellcolor[HTML]{E6A59D}24.81 & 
\cellcolor[HTML]{EAB0A7}31.48 & 
\cellcolor[HTML]{EDBBB1}36.29 \\

\textbf{GPT-5.2} & 
\cellcolor[HTML]{E4A09A}22.22 & 
\cellcolor[HTML]{E4A09A}22.22 & 
\cellcolor[HTML]{E4A09A}22.22 & 
\cellcolor[HTML]{EBB3AA}33.33 & 
\cellcolor[HTML]{EBB3AA}33.33 & 
\cellcolor[HTML]{FAE1D2}50 & 
\cellcolor[HTML]{E5A29B}23.33 & 
\cellcolor[HTML]{F1C5BA}40 & 
\cellcolor[HTML]{F1C5BA}40 & 
\cellcolor[HTML]{E7A7A0}26.29 & 
\cellcolor[HTML]{EAB1A8}31.85 & 
\cellcolor[HTML]{EEBEB4}37.41 \\

\textbf{Llama 3.1 8B} & 
\cellcolor[HTML]{DE8C89}11.11 & 
\cellcolor[HTML]{DE8C89}11.11 & 
\cellcolor[HTML]{DE8C89}11.11 & 
\cellcolor[HTML]{DA7777}0 & 
\cellcolor[HTML]{DA7777}0 & 
\cellcolor[HTML]{DA7777}0 & 
\cellcolor[HTML]{DE8A87}10 & 
\cellcolor[HTML]{DF908C}13.33 & 
\cellcolor[HTML]{E19791}16.67 & 
\cellcolor[HTML]{DD8482}7.04 & 
\cellcolor[HTML]{DD8684}8.15 & 
\cellcolor[HTML]{DE8886}9.26 \\

\textbf{Llama 3.3 70B} & 
\cellcolor[HTML]{DE8C89}11.11 & 
\cellcolor[HTML]{DE8C89}11.11 & 
\cellcolor[HTML]{DE8C89}11.11 & 
\cellcolor[HTML]{DA7777}0 & 
\cellcolor[HTML]{DA7777}0 & 
\cellcolor[HTML]{DA7777}0 & 
\cellcolor[HTML]{DE8A87}10 & 
\cellcolor[HTML]{E7A8A0}26.67 & 
\cellcolor[HTML]{E9ADA5}30 & 
\cellcolor[HTML]{DD8482}7.04 & 
\cellcolor[HTML]{DF8F8B}12.59 & 
\cellcolor[HTML]{DF918D}13.70 \\

\textbf{Mage (DeepSeek-V3)} & \cellcolor[HTML]{E4A09A}22.22 & - & - & \cellcolor[HTML]{FAE1D2}50 & - & - & \cellcolor[HTML]{F1C5BA}40 & - & - & \cellcolor[HTML]{EEBEB4}37.41 & - & - \\

\bottomrule
\end{tabular}%
}
\end{table*}
\begin{table*}[t]
\centering
\renewcommand{\arraystretch}{1.1}
\caption{Test result for Python reference model generation (Red color denotes low pass rate, which green color denotes high)}
\label{tab:python_gen}

\resizebox{0.83\textwidth}{!}{%
\begin{tabular}{lcccccccccccc}
\toprule
\multirow{2}{*}{\textbf{Model/Dataset}} & \multicolumn{3}{c}{\textbf{CPU IP}} & \multicolumn{3}{c}{\textbf{Non-Self-Contained}} & \multicolumn{3}{c}{\textbf{Self-Contained}} & \multicolumn{3}{c}{\textbf{Average}} \\
\cmidrule(lr){2-4} \cmidrule(lr){5-7} \cmidrule(lr){8-10} \cmidrule(lr){11-13}
 & \textbf{pass@1} & \textbf{pass@5} & \textbf{pass@10} & \textbf{pass@1} & \textbf{pass@5} & \textbf{pass@10} & \textbf{pass@1} & \textbf{pass@5} & \textbf{pass@10} & \textbf{pass@1} & \textbf{pass@5} & \textbf{pass@10} \\
\midrule

\textbf{Claude 4.5 Haiku} & 
\cellcolor[HTML]{DA7777}0 & 
\cellcolor[HTML]{DA7777}0 & 
\cellcolor[HTML]{DA7777}0 & 
\cellcolor[HTML]{DA7777}0 & 
\cellcolor[HTML]{DA7777}0 & 
\cellcolor[HTML]{DA7777}0 & 
\cellcolor[HTML]{F1C5BA}40 & 
\cellcolor[HTML]{F1C5BA}40 & 
\cellcolor[HTML]{F4CEC2}43.33 & 
\cellcolor[HTML]{DF908C}13.33 & 
\cellcolor[HTML]{DF908C}13.33 & 
\cellcolor[HTML]{E0928E}14.44 \\

\textbf{Claude 4.5 Opus} & 
\cellcolor[HTML]{DA7777}0 & 
\cellcolor[HTML]{DA7777}0 & 
\cellcolor[HTML]{DA7777}0 & 
\cellcolor[HTML]{DA7777}0 & 
\cellcolor[HTML]{DA7777}0 & 
\cellcolor[HTML]{DA7777}0 & 
\cellcolor[HTML]{F1C5BA}40 & 
\cellcolor[HTML]{F7D7CA}46.67 & 
\cellcolor[HTML]{F7D7CA}46.67 & 
\cellcolor[HTML]{DF908C}13.33 & 
\cellcolor[HTML]{E09590}15.56 & 
\cellcolor[HTML]{E09590}15.56 \\

\textbf{Claude 4.5 Sonnet} & 
\cellcolor[HTML]{DE8C89}11.11 & 
\cellcolor[HTML]{DE8C89}11.11 & 
\cellcolor[HTML]{DE8C89}11.11 & 
\cellcolor[HTML]{DA7777}0 & 
\cellcolor[HTML]{DA7777}0 & 
\cellcolor[HTML]{DE8C89}11.11 & 
\cellcolor[HTML]{EEBCB2}36.67 & 
\cellcolor[HTML]{EEBCB2}36.67 & 
\cellcolor[HTML]{EEBCB2}36.67 & 
\cellcolor[HTML]{E09590}15.93 & 
\cellcolor[HTML]{E09590}15.93 & 
\cellcolor[HTML]{E39C96}19.63 \\

\textbf{DeepSeek-Coder} & 
\cellcolor[HTML]{DA7777}0 & 
\cellcolor[HTML]{DA7777}0 & 
\cellcolor[HTML]{DA7777}0 & 
\cellcolor[HTML]{DA7777}0 & 
\cellcolor[HTML]{DA7777}0 & 
\cellcolor[HTML]{DA7777}0 & 
\cellcolor[HTML]{EEBCB2}36.67 & 
\cellcolor[HTML]{F1C5BA}40 & 
\cellcolor[HTML]{F1C5BA}40 & 
\cellcolor[HTML]{DF8E8A}12.22 & 
\cellcolor[HTML]{DF908C}13.33 & 
\cellcolor[HTML]{DF908C}13.33 \\

\textbf{DeepSeek-R1} & 
\cellcolor[HTML]{DA7777}0 & 
\cellcolor[HTML]{DA7777}0 & 
\cellcolor[HTML]{DA7777}0 & 
\cellcolor[HTML]{DA7777}0 & 
\cellcolor[HTML]{DA7777}0 & 
\cellcolor[HTML]{DA7777}0 & 
\cellcolor[HTML]{F4CEC2}43.33 & 
\cellcolor[HTML]{F4CEC2}43.33 & 
\cellcolor[HTML]{F4CEC2}43.33 & 
\cellcolor[HTML]{E0928E}14.44 & 
\cellcolor[HTML]{E0928E}14.44 & 
\cellcolor[HTML]{E0928E}14.44 \\

\textbf{Gemini 2.5 Pro} & 
\cellcolor[HTML]{DA7777}0 & 
\cellcolor[HTML]{DA7777}0 & 
\cellcolor[HTML]{DA7777}0 & 
\cellcolor[HTML]{DA7777}0 & 
\cellcolor[HTML]{DA7777}0 & 
\cellcolor[HTML]{DA7777}0 & 
\cellcolor[HTML]{F1C5BA}40 & 
\cellcolor[HTML]{F7D7CA}46.67 & 
\cellcolor[HTML]{F7D7CA}46.67 & 
\cellcolor[HTML]{DF908C}13.33 & 
\cellcolor[HTML]{E09590}15.56 & 
\cellcolor[HTML]{E09590}15.56 \\

\textbf{Gemini 3 Flash} & 
\cellcolor[HTML]{DE8C89}11.11 & 
\cellcolor[HTML]{DE8C89}11.11 & 
\cellcolor[HTML]{DE8C89}11.11 & 
\cellcolor[HTML]{DA7777}0 & 
\cellcolor[HTML]{DA7777}0 & 
\cellcolor[HTML]{DA7777}0 & 
\cellcolor[HTML]{EBB3AA}33.33 & 
\cellcolor[HTML]{FAE1D2}50 & 
\cellcolor[HTML]{FAE1D2}50 & 
\cellcolor[HTML]{E0938F}14.81 & 
\cellcolor[HTML]{E39D97}20.37 & 
\cellcolor[HTML]{E39D97}20.37 \\

\textbf{GPT-3.5 Turbo} & 
\cellcolor[HTML]{DA7777}0 & 
\cellcolor[HTML]{DA7777}0 & 
\cellcolor[HTML]{DA7777}0 & 
\cellcolor[HTML]{DA7777}0 & 
\cellcolor[HTML]{DA7777}0 & 
\cellcolor[HTML]{DA7777}0 & 
\cellcolor[HTML]{E19791}16.67 & 
\cellcolor[HTML]{E39C96}20 & 
\cellcolor[HTML]{E5A29B}23.33 & 
\cellcolor[HTML]{DC8180}5.56 & 
\cellcolor[HTML]{DC8381}6.67 & 
\cellcolor[HTML]{DD8683}7.78 \\

\textbf{GPT-4o} & 
\cellcolor[HTML]{DA7777}0 & 
\cellcolor[HTML]{DA7777}0 & 
\cellcolor[HTML]{DA7777}0 & 
\cellcolor[HTML]{DA7777}0 & 
\cellcolor[HTML]{DA7777}0 & 
\cellcolor[HTML]{DA7777}0 & 
\cellcolor[HTML]{EBB3AA}33.33 & 
\cellcolor[HTML]{F1C5BA}40 & 
\cellcolor[HTML]{F1C5BA}40 & 
\cellcolor[HTML]{DE8C89}11.11 & 
\cellcolor[HTML]{DF908C}13.33 & 
\cellcolor[HTML]{DF908C}13.33 \\

\textbf{GPT-5} & 
\cellcolor[HTML]{DA7777}0 & 
\cellcolor[HTML]{DA7777}0 & 
\cellcolor[HTML]{DA7777}0 & 
\cellcolor[HTML]{DA7777}0 & 
\cellcolor[HTML]{DE8C89}11.11 & 
\cellcolor[HTML]{DE8C89}11.11 & 
\cellcolor[HTML]{F1C5BA}40 & 
\cellcolor[HTML]{F4CEC2}43.33 & 
\cellcolor[HTML]{F4CEC2}43.33 & 
\cellcolor[HTML]{DF908C}13.33 & 
\cellcolor[HTML]{E29994}18.15 & 
\cellcolor[HTML]{E29994}18.15 \\

\textbf{GPT-5.2} & 
\cellcolor[HTML]{DA7777}0 & 
\cellcolor[HTML]{DA7777}0 & 
\cellcolor[HTML]{DA7777}0 & 
\cellcolor[HTML]{DA7777}0 & 
\cellcolor[HTML]{DE8C89}11.11 & 
\cellcolor[HTML]{DE8C89}11.11 & 
\cellcolor[HTML]{F4CEC2}43.33 & 
\cellcolor[HTML]{F4CEC2}43.33 & 
\cellcolor[HTML]{F4CEC2}43.33 & 
\cellcolor[HTML]{E0928E}14.44 & 
\cellcolor[HTML]{E29994}18.15 & 
\cellcolor[HTML]{E29994}18.15 \\

\textbf{Llama 3.1 8B} & 
\cellcolor[HTML]{DA7777}0 & 
\cellcolor[HTML]{DA7777}0 & 
\cellcolor[HTML]{DA7777}0 & 
\cellcolor[HTML]{DA7777}0 & 
\cellcolor[HTML]{DA7777}0 & 
\cellcolor[HTML]{DA7777}0 & 
\cellcolor[HTML]{DC8381}6.67 & 
\cellcolor[HTML]{E19791}16.67 & 
\cellcolor[HTML]{E39C96}20 & 
\cellcolor[HTML]{DA7B7A}2.22 & 
\cellcolor[HTML]{DC8180}5.56 & 
\cellcolor[HTML]{DC8381}6.67 \\

\textbf{Llama 3.3 70B} & 
\cellcolor[HTML]{DA7777}0 & 
\cellcolor[HTML]{DA7777}0 & 
\cellcolor[HTML]{DA7777}0 & 
\cellcolor[HTML]{DA7777}0 & 
\cellcolor[HTML]{DA7777}0 & 
\cellcolor[HTML]{DA7777}0 & 
\cellcolor[HTML]{E7A8A0}26.67 & 
\cellcolor[HTML]{E9ADA5}30 & 
\cellcolor[HTML]{E9ADA5}30 & 
\cellcolor[HTML]{DD8885}8.89 & 
\cellcolor[HTML]{DE8A87}10 & 
\cellcolor[HTML]{DE8A87}10 \\

\bottomrule
\end{tabular}%
}
\end{table*}
\begin{table*}[t]
\centering
\renewcommand{\arraystretch}{1.2}
\caption{Test results for Verilog debugging (Red color denotes low pass rate, which green color denotes high)}
\label{tab:debug_zero_shot}

\resizebox{0.99\textwidth}{!}{%
\begin{tabular}{lccccccccccccccc}
\toprule
\multirow{2}{*}{\textbf{Model/Dataset}} & \multicolumn{3}{c}{\textbf{Arithmetic Bug}} & \multicolumn{3}{c}{\textbf{Assignment Bug}} & \multicolumn{3}{c}{\textbf{State Machine Bug}} & \multicolumn{3}{c}{\textbf{Timing Bug}} & \multicolumn{3}{c}{\textbf{Average}} \\
\cmidrule(lr){2-4} \cmidrule(lr){5-7} \cmidrule(lr){8-10} \cmidrule(lr){11-13} \cmidrule(lr){14-16}
 & \textbf{pass@1} & \textbf{pass@5} & \textbf{pass@10} & \textbf{pass@1} & \textbf{pass@5} & \textbf{pass@10} & \textbf{pass@1} & \textbf{pass@5} & \textbf{pass@10} & \textbf{pass@1} & \textbf{pass@5} & \textbf{pass@10} & \textbf{pass@1} & \textbf{pass@5} & \textbf{pass@10} \\
\midrule

\textbf{Claude 4.5 Haiku} & 
\cellcolor[HTML]{F2CABE}41.67 & 
\cellcolor[HTML]{D2E4CA}62.50 & 
\cellcolor[HTML]{C6E4C6}66.67 & 
\cellcolor[HTML]{FAE1D2}50 & 
\cellcolor[HTML]{E5E3CD}56.67 & 
\cellcolor[HTML]{E5E3CD}56.67 & 
\cellcolor[HTML]{E19791}16.67 & 
\cellcolor[HTML]{E19791}16.67 & 
\cellcolor[HTML]{E19791}16.67 & 
\cellcolor[HTML]{E9AFA7}31.03 & 
\cellcolor[HTML]{F2C9BD}41.38 & 
\cellcolor[HTML]{F2C9BD}41.38 & 
\cellcolor[HTML]{ECB7AE}34.84 & 
\cellcolor[HTML]{F4D1C4}44.30 & 
\cellcolor[HTML]{F5D4C7}45.35 \\

\textbf{Claude 4.5 Opus} & 
\cellcolor[HTML]{DFE3CC}58.33 & 
\cellcolor[HTML]{AED1AE}75 & 
\cellcolor[HTML]{A2C8A2}79.17 & 
\cellcolor[HTML]{EFE2CF}53.33 & 
\cellcolor[HTML]{DAE4CB}60 & 
\cellcolor[HTML]{BCDCBC}70 & 
\cellcolor[HTML]{EBB3AA}33.33 & 
\cellcolor[HTML]{FAE1D2}50 & 
\cellcolor[HTML]{C6E4C6}66.67 & 
\cellcolor[HTML]{F5D3C5}44.83 & 
\cellcolor[HTML]{E9E2CE}55.17 & 
\cellcolor[HTML]{D4E4CA}62.07 & 
\cellcolor[HTML]{F7DACC}47.45 & 
\cellcolor[HTML]{DAE4CB}60.04 & 
\cellcolor[HTML]{BDDEBD}69.48 \\

\textbf{Claude 4.5 Sonnet} & 
\cellcolor[HTML]{DB7F7D}4.17 & 
\cellcolor[HTML]{DF8F8B}12.50 & 
\cellcolor[HTML]{DF8F8B}12.50 & 
\cellcolor[HTML]{DC8381}6.67 & 
\cellcolor[HTML]{DC8381}6.67 & 
\cellcolor[HTML]{DE8A87}10 & 
\cellcolor[HTML]{DA7777}0 & 
\cellcolor[HTML]{DA7777}0 & 
\cellcolor[HTML]{DA7777}0 & 
\cellcolor[HTML]{DA7777}0 & 
\cellcolor[HTML]{DA7777}0 & 
\cellcolor[HTML]{DA7777}0 & 
\cellcolor[HTML]{DB7C7B}2.71 & 
\cellcolor[HTML]{DC807E}4.79 & 
\cellcolor[HTML]{DC8180}5.62 \\

\textbf{DeepSeek-Coder-V2.5} & 
\cellcolor[HTML]{DFE3CC}58.33 & 
\cellcolor[HTML]{D2E4CA}62.50 & 
\cellcolor[HTML]{D2E4CA}62.50 & 
\cellcolor[HTML]{EFE2CF}53.33 & 
\cellcolor[HTML]{EFE2CF}53.33 & 
\cellcolor[HTML]{EFE2CF}53.33 & 
\cellcolor[HTML]{DA7777}0 & 
\cellcolor[HTML]{E19791}16.67 & 
\cellcolor[HTML]{E19791}16.67 & 
\cellcolor[HTML]{E9AFA7}31.03 & 
\cellcolor[HTML]{EFC0B5}37.93 & 
\cellcolor[HTML]{EFC0B5}37.93 & 
\cellcolor[HTML]{EDBAB0}35.67 & 
\cellcolor[HTML]{F3CCC0}42.61 & 
\cellcolor[HTML]{F3CCC0}42.61 \\

\textbf{DeepSeek-R1} & 
\cellcolor[HTML]{F6D5C8}45.83 & 
\cellcolor[HTML]{AED1AE}75 & 
\cellcolor[HTML]{A2C8A2}79.17 & 
\cellcolor[HTML]{FAE1D2}50 & 
\cellcolor[HTML]{C6E4C6}66.67 & 
\cellcolor[HTML]{B2D5B2}73.33 & 
\cellcolor[HTML]{FAE1D2}50 & 
\cellcolor[HTML]{FAE1D2}50 & 
\cellcolor[HTML]{C6E4C6}66.67 & 
\cellcolor[HTML]{E9AFA7}31.03 & 
\cellcolor[HTML]{E9E2CE}55.17 & 
\cellcolor[HTML]{BFDFBF}68.97 & 
\cellcolor[HTML]{F4D1C4}44.22 & 
\cellcolor[HTML]{D5E4CA}61.71 & 
\cellcolor[HTML]{B6D8B6}72.03 \\

\textbf{Gemini 2.5 Pro} & 
\cellcolor[HTML]{DA7777}0 & 
\cellcolor[HTML]{DA7777}0 & 
\cellcolor[HTML]{DA7777}0 & 
\cellcolor[HTML]{DB7D7C}3.33 & 
\cellcolor[HTML]{DB7D7C}3.33 & 
\cellcolor[HTML]{DB7D7C}3.33 & 
\cellcolor[HTML]{DA7777}0 & 
\cellcolor[HTML]{DA7777}0 & 
\cellcolor[HTML]{DA7777}0 & 
\cellcolor[HTML]{DA7777}0 & 
\cellcolor[HTML]{DA7777}0 & 
\cellcolor[HTML]{DA7777}0 & 
\cellcolor[HTML]{DA7878}0.83 & 
\cellcolor[HTML]{DA7878}0.83 & 
\cellcolor[HTML]{DA7878}0.83 \\

\textbf{Gemini 3 Flash} & 
\cellcolor[HTML]{EEBFB4}37.50 & 
\cellcolor[HTML]{D2E4CA}62.50 & 
\cellcolor[HTML]{C6E4C6}66.67 & 
\cellcolor[HTML]{EEBCB2}36.67 & 
\cellcolor[HTML]{EFE2CF}53.33 & 
\cellcolor[HTML]{EFE2CF}53.33 & 
\cellcolor[HTML]{E19791}16.67 & 
\cellcolor[HTML]{FAE1D2}50 & 
\cellcolor[HTML]{FAE1D2}50 & 
\cellcolor[HTML]{E7A9A2}27.59 & 
\cellcolor[HTML]{F2C9BD}41.38 & 
\cellcolor[HTML]{F2C9BD}41.38 & 
\cellcolor[HTML]{E9ADA5}29.61 & 
\cellcolor[HTML]{F4E1D0}51.80 & 
\cellcolor[HTML]{F1E1D0}52.84 \\

\textbf{GPT-3.5 Turbo} & 
\cellcolor[HTML]{E19791}16.67 & 
\cellcolor[HTML]{E6A59E}25 & 
\cellcolor[HTML]{E6A59E}25 & 
\cellcolor[HTML]{DC8381}6.67 & 
\cellcolor[HTML]{E7A8A0}26.67 & 
\cellcolor[HTML]{F7D7CA}46.67 & 
\cellcolor[HTML]{E19791}16.67 & 
\cellcolor[HTML]{EBB3AA}33.33 & 
\cellcolor[HTML]{EBB3AA}33.33 & 
\cellcolor[HTML]{DA7777}0 & 
\cellcolor[HTML]{DA7777}0 & 
\cellcolor[HTML]{DA7777}0 & 
\cellcolor[HTML]{DE8A87}10.00 & 
\cellcolor[HTML]{E49E98}21.25 & 
\cellcolor[HTML]{E7A7A0}26.25 \\

\textbf{GPT-4o} & 
\cellcolor[HTML]{E19791}16.67 & 
\cellcolor[HTML]{E8ACA4}29.17 & 
\cellcolor[HTML]{EBB3AA}33.33 & 
\cellcolor[HTML]{DC8381}6.67 & 
\cellcolor[HTML]{EEBCB2}36.67 & 
\cellcolor[HTML]{F1C5BA}40 & 
\cellcolor[HTML]{DA7777}0 & 
\cellcolor[HTML]{DA7777}0 & 
\cellcolor[HTML]{DA7777}0 & 
\cellcolor[HTML]{DD8482}6.90 & 
\cellcolor[HTML]{E19892}17.24 & 
\cellcolor[HTML]{E39E97}20.69 & 
\cellcolor[HTML]{DD8583}7.56 & 
\cellcolor[HTML]{E39E98}20.77 & 
\cellcolor[HTML]{E5A29C}23.50 \\

\textbf{GPT-5} & 
\cellcolor[HTML]{FAE1D2}50 & 
\cellcolor[HTML]{C6E4C6}66.67 & 
\cellcolor[HTML]{BADBBA}70.83 & 
\cellcolor[HTML]{F4CEC2}43.33 & 
\cellcolor[HTML]{DAE4CB}60 & 
\cellcolor[HTML]{D0E5C9}63.33 & 
\cellcolor[HTML]{FAE1D2}50 & 
\cellcolor[HTML]{FAE1D2}50 & 
\cellcolor[HTML]{C6E4C6}66.67 & 
\cellcolor[HTML]{E9AFA7}31.03 & 
\cellcolor[HTML]{F4E1D0}51.72 & 
\cellcolor[HTML]{E9E2CE}55.17 & 
\cellcolor[HTML]{F4CFC2}43.59 & 
\cellcolor[HTML]{E3E3CD}57.10 & 
\cellcolor[HTML]{CEE5C9}64 \\

\textbf{GPT-5.2} & 
\cellcolor[HTML]{F6D5C8}45.83 & 
\cellcolor[HTML]{D2E4CA}62.50 & 
\cellcolor[HTML]{C6E4C6}66.67 & 
\cellcolor[HTML]{F1C5BA}40 & 
\cellcolor[HTML]{DAE4CB}60 & 
\cellcolor[HTML]{C6E4C6}66.67 & 
\cellcolor[HTML]{DA7777}0 & 
\cellcolor[HTML]{FAE1D2}50 & 
\cellcolor[HTML]{FAE1D2}50 & 
\cellcolor[HTML]{F5D3C5}44.83 & 
\cellcolor[HTML]{F4E1D0}51.72 & 
\cellcolor[HTML]{E9E2CE}55.17 & 
\cellcolor[HTML]{EAB2A9}32.66 & 
\cellcolor[HTML]{E7E2CE}56.05 & 
\cellcolor[HTML]{DBE4CB}59.63 \\

\textbf{Llama 3.1 8B} & 
\cellcolor[HTML]{DB7F7D}4.17 & 
\cellcolor[HTML]{DD8784}8.33 & 
\cellcolor[HTML]{E39E98}20.83 & 
\cellcolor[HTML]{DB7D7C}3.33 & 
\cellcolor[HTML]{E5A29B}23.33 & 
\cellcolor[HTML]{EEBCB2}36.67 & 
\cellcolor[HTML]{E19791}16.67 & 
\cellcolor[HTML]{E19791}16.67 & 
\cellcolor[HTML]{E19791}16.67 & 
\cellcolor[HTML]{DD8482}6.90 & 
\cellcolor[HTML]{DE8B87}10.34 & 
\cellcolor[HTML]{E19892}17.24 & 
\cellcolor[HTML]{DD8683}7.77 & 
\cellcolor[HTML]{E0938E}14.67 & 
\cellcolor[HTML]{E5A19B}22.85 \\

\textbf{Llama 3.3 70B} & 
\cellcolor[HTML]{EEBFB4}37.50 & 
\cellcolor[HTML]{EEBFB4}37.50 & 
\cellcolor[HTML]{F2CABE}41.67 & 
\cellcolor[HTML]{E39C96}20 & 
\cellcolor[HTML]{EBB3AA}33.33 & 
\cellcolor[HTML]{EEBCB2}36.67 & 
\cellcolor[HTML]{E19791}16.67 & 
\cellcolor[HTML]{E19791}16.67 & 
\cellcolor[HTML]{E19791}16.67 & 
\cellcolor[HTML]{E39E97}20.69 & 
\cellcolor[HTML]{E9AFA7}31.03 & 
\cellcolor[HTML]{ECB7AD}34.48 & 
\cellcolor[HTML]{E5A39C}23.71 & 
\cellcolor[HTML]{E9ADA5}29.63 & 
\cellcolor[HTML]{EAB1A9}32.37 \\

\bottomrule
\end{tabular}%
}
\end{table*}

\begin{figure}[t]
    \centering
    \includegraphics[width=0.97\columnwidth]{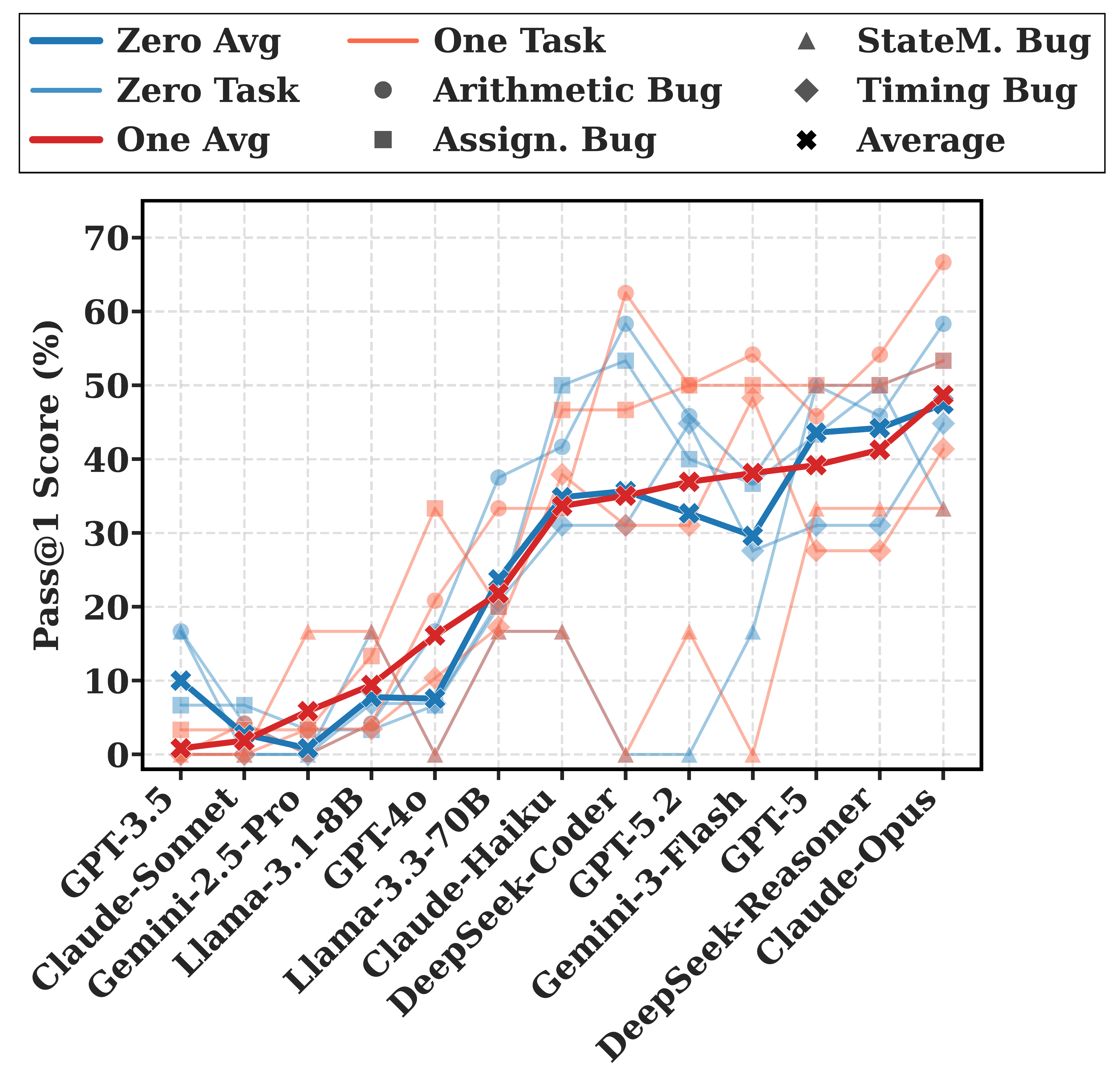}
    \caption{Performance comparison between Zero-Shot debug tasks and One-Shot debug tasks. In Zero-Shot tests, LLMs are only aware of the buggy code and the existence of bugs, while in One-Shot tests, LLMs have access to the complete waveform file.}
    \label{fig:one-shot debugging}
\end{figure}

\begin{figure}[t]
    \centering
    \includegraphics[width=0.97\linewidth]{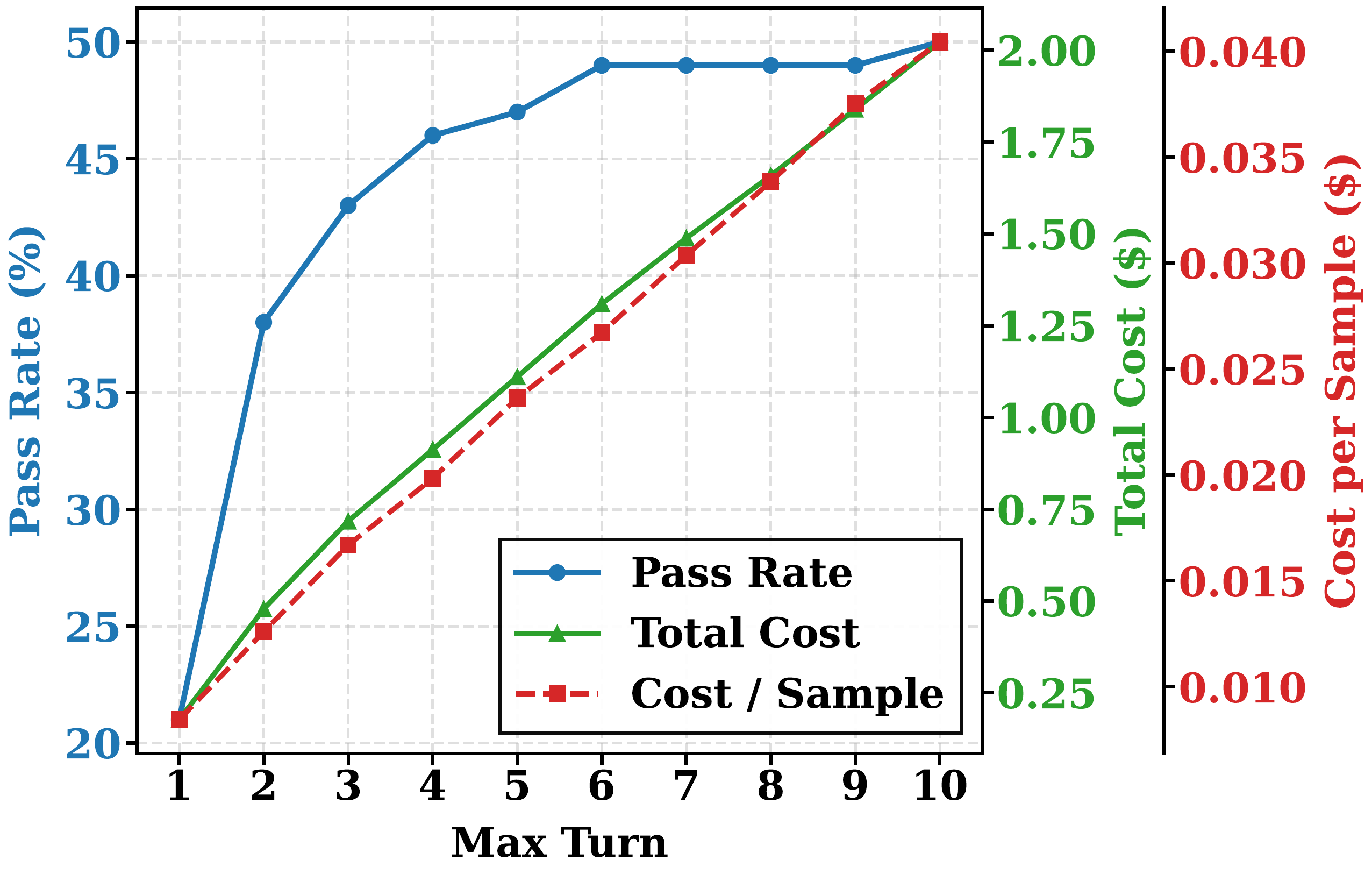} 
    \caption{Performance and cost efficiency analysis of DeepSeek Reasoner across multiple turns. The blue line represents the cumulative pass rate, while the red dashed line indicates the average cost incurred per solved instance.}
    \label{fig:training_gen}
\end{figure}

\subsection{Primary Results}
We present the results for Verilog generation in \autoref{tab:verilog_gen}, 
Verilog debugging in \autoref{tab:debug_zero_shot}, and reference model 
generation in \autoref{tab:python_gen}. 
Our key observations are as follows:

\textbf{Significant performance drop compared to existing benchmarks.}
Despite testing on advanced models, the average pass@1 is relatively low. 
For instance, GPT-4o achieves $\sim$65\% pass@1 on RTLLM V2 and VerilogEval V2, 
but only 10.37\% on our dataset. Even MAGE, which exceeds 95\% accuracy on 
VerilogEval, achieves only 37.41\% on \xname, though it remains the top performer. 
These results underscore the difficulty of our dataset and demonstrate that 
\xname~serves as a rigorous next-generation benchmark for AI-aided chip design. 
Notably, CPU IP design proves to be the most challenging task, with no model 
exceeding 22.22\% accuracy, indicating that current LLMs require further 
optimization for complex Verilog generation.

\textbf{Debugging capability exceeds generation capability.}
LLMs consistently perform better on Verilog debugging than generation when given the same Verilog module. For example, Claude-opus achieves 47.45\% on debugging 
compared to 30.74\% on generation. This performance gap highlights the potential 
of LLMs in debugging tasks and suggests that multi-turn self-correction workflows 
could significantly enhance generation accuracy.

\textbf{Python generation outperforms Verilog on simple modules, but underperforms on complex modules.}
As shown in \autoref{tab:python_gen}, for simple self-contained modules, LLMs achieve 5\%--20\% higher accuracy on Python reference model generation compared to Verilog generation, likely due to Python's prevalence in LLM pre-training data.
Although the accuracy is higher than Verilog generation, the absolute pass rate remains below 50\%, far from the reliable golden reference models needed to validate Verilog code.
This advantage further diminishes as module complexity increases: pass@10 rates drop to below 11.11\% for complex non-self-contained modules and CPU IP designs.
This reveals that while LLMs possess strong Python syntax knowledge, they lack hardware-specific modeling capabilities.
By improving these capabilities, we have the potential to significantly enhance LLMs' impact in the chip design industry.

\subsection{One-shot debugging Analysis}
To evaluate whether LLM can utilize waveform data for debugging like human engineers, we compare zero-shot and one-shot debugging in ~\autoref{fig:one-shot debugging}. 
In zero-shot debugging, LLM receives only the module description and the buggy implementation. 
In one-shot debugging, we additionally provide the waveform file (.vcd) in the 
prompt, mimicking real-world workflows where engineers analyze signals to 
detect mismatches. 
Our results show mixed performance: one-shot debugging outperforms zero-shot on 8 models but underperforms on 13. This suggests that while waveform data contains critical debugging information, most current LLMs lack the capability to interpret it effectively. 
Consequently, we posit that developing agentic tools for waveform analysis and training LLMs specifically on waveform-aware debugging are promising future research directions.

\subsection{Reference Model Training Dataset Generation}
We evaluated our reference model training dataset generation tool using DeepSeek V3.2 on 100 Verilog samples from the QiMeng CodeV-R1 training dataset.
We tested the pass rate and cost under different maximum debugging iterations.
As shown in \autoref{fig:training_gen}, initially, the pass rate increases with maximum iteration count, reaching 50
Beyond this point, the model reaches its debugging limitation and the pass rate plateaus.
This means that with multi-turn debugging, our tool can generate approximately n/2 Python reference models from n Verilog training samples.

While higher iteration counts improve pass rates, they also increase api costs. Both the total cost and average cost per successful sample continue to rise with maximum iteration count.
This suggests a trade-off: we should use lower iteration counts when Verilog data is abundant to minimize costs, and higher iteration counts when data is limited to maximize the number of generated reference models.

We further validated the scalability of our tool by processing 10,000 samples with a maximum iteration count of 1, successfully generating 2,206 verified Python reference models, demonstrating the effectiveness of our proposed tool.

\subsection{Cost Analysis}
We detail token consumption and associated costs in ~\autoref{tab:cost_verilog_gen} to help researchers estimate budgets when evaluating their LLMs and agentic frameworks on ~\xname{}.
Among all models, DeepSeek V3.2-coder emerges as the most cost-efficient model for Verilog generation, offering the lowest cost per pass@1 while maintaining relatively high code quality.
Although Claude-4.5-opus achieves slightly higher pass@1, the high cost of the Claude API makes its Cost/Pass@1 over 275$\times$ higher.

\begin{table}[t]
\centering
\renewcommand{\arraystretch}{1.2}
\caption{Cost Efficiency Analysis (Verilog Pass@1). Cost calculations are based on standard API pricing per million tokens.}
\label{tab:cost_verilog_gen}

\resizebox{\columnwidth}{!}{%
\begin{tabular}{lrrrrrr}
\toprule
\textbf{Model} & 
\makecell[r]{\textbf{Input} \\ (Tokens)} & 
\makecell[r]{\textbf{Output} \\ (Tokens)} & 
\makecell[r]{\textbf{Cost} \\ (\$)} & 
\makecell[r]{\textbf{Pass@1} \\ (\%)} & 
\makecell[r]{\textbf{Cost/Pass@1} \\ (\$)} \\
\midrule
\textbf{Claude 4.5 Haiku} & 25,076 & 24,925 & 0.037 & 19.26 & 0.194 \\
\textbf{Claude 4.5 Opus} & 25,076 & 32,221 & 2.793 & 30.74 & 9.085 \\
\textbf{Claude 4.5 Sonnet} & 25,076 & 29,479 & 0.517 & 12.59 & 4.109 \\
\textbf{DeepSeek-Coder} & 21,933 & 23,992 & 0.010 & 29.26 & 0.033 \\
\textbf{DeepSeek-R1} & 21,933 & 171,740 & 0.388 & 23.70 & 1.638 \\
\textbf{Gemini 2.5 Pro} & 23,409 & 299,064 & 3.222 & 1.11 & 290.283 \\
\textbf{Gemini 3 Flash} & 23,409 & 418,835 & 0.127 & 30.74 & 0.414 \\
\textbf{GPT-3.5 Turbo} & 21,710 & 14,030 & 0.032 & 5.56 & 0.574 \\
\textbf{GPT-4o} & 21,699 & 19,375 & 0.248 & 10.37 & 2.391 \\
\textbf{GPT-5} & 21,654 & 152,233 & 9.459 & 24.81 & 38.120 \\
\textbf{GPT-5.2} & 21,654 & 24,021 & 1.766 & 26.29 & 6.718 \\
\textbf{Llama 3.1 8B} & 22,699 & 32,271 & 0.003 & 7.04 & 0.039 \\
\textbf{Llama 3.3 70B} & 22,699 & 16,624 & 0.012 & 7.04 & 0.169 \\
\bottomrule
\end{tabular}%
}
\end{table}

\section{Take-Aways}
Based on our evaluation, we identify several promising future research directions:
\begin{itemize}
\item \textbf{Hierarchical Verilog Generation.} The success rate for generating only the top module is already below 22.22\% and drops to 0\% when generating complete hierarchical designs, highlighting the critical need for hierarchical design capabilities that align with industrial practices.
\item \textbf{Waveform-Aware Debugging.} While LLMs demonstrate stronger abilities in debugging than in Verilog generation, our experiments in~\autoref{fig:one-shot debugging} reveal poor waveform interpretation capabilities. Since waveforms are crucial debugging information sources in industry, enhancing LLMs' waveform analysis could significantly improve their practical utility.
\item \textbf{Reference Model Generation.} Despite LLMs' proficiency in other Python tasks, their pass rates for Python reference model generation are extremely low. This gap indicates that while LLMs understand Python syntax, they lack hardware behavior modeling knowledge, representing a significant unexplored opportunity for both chip design and verification workflows.
\end{itemize}

\section{Related Work}

\textbf{Verilog Generation Benchmark.}
Early stage benchmarks VerilogEval~\cite{liu2023verilogeval}, proposed in 2023, pioneered LLM evaluation for Verilog code completion, while successive work, VerilogEval V2~\cite{pinckney2025revisiting}, introduced robust infrastructure for specification-to-RTL tasks. 
Contemporaneous benchmarks RTLLM~\cite{lu2024rtllm} and RTLLM V2~\cite{liu2024openllm} also focus on the same spec-RTL tasks and gained popularity. 
However, these benchmarks have key limitations compared to \xname{}: they exclusively target code generation while neglecting debugging and reference model generation, and their test cases present low complexity for current LLMs, with VerilogEval V2 containing only self-contained modules under 76 lines.

Recent ArXiv-released benchmarks in late 2025 address complexity differently. 
RealBench~\cite{jin2025realbench} features open-source IPs with extremely high complexity and high-quality specifications, but focuses only on generation and proves too difficult for current LLMs. CVDP~\cite{guo2025comprehensive} expands to 13 task categories, including code modification and code improvement, but lacks comprehensive reference model evaluation and offers limited debugging cases.
\xname{}, as a contemporary work, provides essential complementary capabilities with a specific emphasis on debugging and reference model generation overlooked by prior work.
We offer distinctly different test cases for Verilog generation, and the combined usage of our benchmark with these contemporary works can provide a more comprehensive evaluation of LLMs' capabilities in chip design.
Furthermore, we also provide a specialized toolbox to accelerate future research about reference model generation.

\textbf{LLM for testbench generation.}
Recent works on LLM-assisted Verilog testbench generation, like AutoBench~\cite{qiu2024autobench}, CorrectBench~\cite{qiu2025correctbench}, and PRO-V~\cite{zhao2025pro}, also consider reference model generation as a sub-task.
However, they only evaluate the correctness of the entire testbench and neglect the correctness of the reference model itself.
Although they use 10-sampling and majority voting to increase accuracy, our evaluation shows that the pass@10 of the most advanced model is only around 20\%, meaning they are highly likely to add stimulus signals to an incorrect reference model.
Therefore, we believe that testbench generation could benefit significantly from accurate reference models, which further demonstrates the value of our benchmark in evaluating reference model generation capabilities.
\section{Conclusion}
This paper introduces \xname{}, a comprehensive benchmark that addresses critical gaps in evaluating LLMs for chip design applications.
By expanding beyond simple Verilog generation to include complex modules, debugging tasks, and reference model generation, \xname{} provides a more realistic assessment framework that better reflects industrial deployment requirements.
Evaluation of various models on \xname{} reveals that current models have significant limitations in AI-aided chip design and remain far from ready for real industrial workflow integration.
We highlight the importance of debugging and reference model generation capabilities and provide a toolbox to support future research in reference model generation.


\section*{Impact Statement}

This paper presents work whose goal is to advance the field of Machine
Learning. There are many potential societal consequences of our work, none
which we feel must be specifically highlighted here.

\nocite{langley00}
\bibliography{example_paper}

@inproceedings{qiu2025correctbench,
  title={Correctbench: Automatic testbench generation with functional self-correction using llms for hdl design},
  author={Qiu, Ruidi and Zhang, Grace Li and Drechsler, Rolf and Schlichtmann, Ulf and Li, Bing},
  booktitle={2025 Design, Automation \& Test in Europe Conference (DATE)},
  pages={1--7},
  year={2025},
  organization={IEEE}
}

@inproceedings{qiu2024autobench,
  title={Autobench: Automatic testbench generation and evaluation using llms for hdl design},
  author={Qiu, Ruidi and Zhang, Grace Li and Drechsler, Rolf and Schlichtmann, Ulf and Li, Bing},
  booktitle={Proceedings of the 2024 ACM/IEEE International Symposium on Machine Learning for CAD},
  pages={1--10},
  year={2024}
}

@article{thakur2024verigen,
  title={Verigen: A large language model for verilog code generation},
  author={Thakur, Shailja and Ahmad, Baleegh and Pearce, Hammond and Tan, Benjamin and Dolan-Gavitt, Brendan and Karri, Ramesh and Garg, Siddharth},
  journal={ACM Transactions on Design Automation of Electronic Systems},
  volume={29},
  number={3},
  pages={1--31},
  year={2024},
  publisher={ACM New York, NY}
}

@inproceedings{nadimi2025pyranet,
  title={Pyranet: A multi-layered hierarchical dataset for verilog},
  author={Nadimi, Bardia and Boutaib, Ghali Omar and Zheng, Hao},
  booktitle={2025 62nd ACM/IEEE Design Automation Conference (DAC)},
  pages={1--7},
  year={2025},
  organization={IEEE}
}

@inproceedings{zhu2025qimeng,
  title={QiMeng-CodeV-R1: Reasoning-Enhanced Verilog Generation},
  author={Zhu, Yaoyu and Huang, Di and Lyu, Hanqi and Zhang, Xiaoyun and Li, Chongxiao and Shi, Wenxuan and Wu, Yutong and Mu, Jianan and Wang, Jinghua and Jin, Pengwei and others},
  booktitle={The Thirty-ninth Annual Conference on Neural Information Processing Systems},
  year={2025}
}

@article{li2025large,
  title={Large language model-aware in-context learning for code generation},
  author={Li, Jia and Tao, Chongyang and Li, Jia and Li, Ge and Jin, Zhi and Zhang, Huangzhao and Fang, Zheng and Liu, Fang},
  journal={ACM Transactions on Software Engineering and Methodology},
  volume={34},
  number={7},
  pages={1--33},
  year={2025},
  publisher={ACM New York, NY}
}

@inproceedings{zhong2024can,
  title={Can llm replace stack overflow? a study on robustness and reliability of large language model code generation},
  author={Zhong, Li and Wang, Zilong},
  booktitle={Proceedings of the AAAI conference on artificial intelligence},
  volume={38},
  number={19},
  pages={21841--21849},
  year={2024}
}

@article{liu2023your,
  title={Is your code generated by chatgpt really correct? rigorous evaluation of large language models for code generation},
  author={Liu, Jiawei and Xia, Chunqiu Steven and Wang, Yuyao and Zhang, Lingming},
  journal={Advances in Neural Information Processing Systems},
  volume={36},
  pages={21558--21572},
  year={2023}
}

@inproceedings{du2024evaluating,
  title={Evaluating large language models in class-level code generation},
  author={Du, Xueying and Liu, Mingwei and Wang, Kaixin and Wang, Hanlin and Liu, Junwei and Chen, Yixuan and Feng, Jiayi and Sha, Chaofeng and Peng, Xin and Lou, Yiling},
  booktitle={Proceedings of the IEEE/ACM 46th International Conference on Software Engineering},
  pages={1--13},
  year={2024}
}

@article{gu2025effectiveness,
  title={On the effectiveness of large language models in domain-specific code generation},
  author={Gu, Xiaodong and Chen, Meng and Lin, Yalan and Hu, Yuhan and Zhang, Hongyu and Wan, Chengcheng and Wei, Zhao and Xu, Yong and Wang, Juhong},
  journal={ACM Transactions on Software Engineering and Methodology},
  volume={34},
  number={3},
  pages={1--22},
  year={2025},
  publisher={ACM New York, NY}
}

@inproceedings{nadimi2024multi,
  title={A multi-expert large language model architecture for verilog code generation},
  author={Nadimi, Bardia and Zheng, Hao},
  booktitle={2024 IEEE LLM Aided Design Workshop (LAD)},
  pages={1--5},
  year={2024},
  organization={IEEE}
}

@inproceedings{wang2023codet5+,
  title={Codet5+: Open code large language models for code understanding and generation},
  author={Wang, Yue and Le, Hung and Gotmare, Akhilesh and Bui, Nghi and Li, Junnan and Hoi, Steven},
  booktitle={Proceedings of the 2023 conference on empirical methods in natural language processing},
  pages={1069--1088},
  year={2023}
}

@inproceedings{ross2023programmer,
  title={The programmer’s assistant: Conversational interaction with a large language model for software development},
  author={Ross, Steven I and Martinez, Fernando and Houde, Stephanie and Muller, Michael and Weisz, Justin D},
  booktitle={Proceedings of the 28th International Conference on Intelligent User Interfaces},
  pages={491--514},
  year={2023}
}

@inproceedings{cai2024advancing,
  title={Advancing knowledge together: integrating large language model-based conversational AI in small group collaborative learning},
  author={Cai, Zhenyao and Park, Seehee and Nixon, Nia and Doroudi, Shayan},
  booktitle={Extended Abstracts of the CHI Conference on Human Factors in Computing Systems},
  pages={1--9},
  year={2024}
}

@article{wang2025ai,
  title={Ai agentic programming: A survey of techniques, challenges, and opportunities},
  author={Wang, Huanting and Gong, Jingzhi and Zhang, Huawei and Xu, Jie and Wang, Zheng},
  journal={arXiv preprint arXiv:2508.11126},
  year={2025}
}

@article{dong2025survey,
  title={A survey on code generation with llm-based agents},
  author={Dong, Yihong and Jiang, Xue and Qian, Jiaru and Wang, Tian and Zhang, Kechi and Jin, Zhi and Li, Ge},
  journal={arXiv preprint arXiv:2508.00083},
  year={2025}
}

@article{guo2025comprehensive,
  title={A Comprehensive Survey on Benchmarks and Solutions in Software Engineering of LLM-Empowered Agentic System},
  author={Guo, Jiale and Huang, Suizhi and Li, Mei and Huang, Dong and Chen, Xingsheng and Zhang, Regina and Guo, Zhijiang and Yu, Han and Yiu, Siu-Ming and Lio, Pietro and others},
  journal={arXiv preprint arXiv:2510.09721},
  year={2025}
}

@article{ishibashi2024self,
  title={Self-organized agents: A llm multi-agent framework toward ultra large-scale code generation and optimization},
  author={Ishibashi, Yoichi and Nishimura, Yoshimasa},
  journal={arXiv preprint arXiv:2404.02183},
  year={2024}
}

@article{islam2024mapcoder,
  title={Mapcoder: Multi-agent code generation for competitive problem solving},
  author={Islam, Md Ashraful and Ali, Mohammed Eunus and Parvez, Md Rizwan},
  journal={arXiv preprint arXiv:2405.11403},
  year={2024}
}

@article{swaroopa2024evaluating,
  title={Evaluating large language models for automatic register transfer logic generation via high-level synthesis},
  author={Swaroopa, Sneha and Mukherjee, Rijoy and Debnath, Anushka and Chakraborty, Rajat Subhra},
  journal={arXiv preprint arXiv:2408.02793},
  year={2024}
}

@article{abdollahi2024hardware,
  title={Hardware design and verification with large language models: A scoping review, challenges, and open issues},
  author={Abdollahi, Meisam and Yeganli, Seyedeh Faegheh and Baharloo, Mohammad and Baniasadi, Amirali},
  journal={Electronics},
  volume={14},
  number={1},
  pages={120},
  year={2024},
  publisher={MDPI}
}

@inproceedings{gai2025exploring,
  title={Exploring code language models for automated hls-based hardware generation: Benchmark, infrastructure and analysis},
  author={Gai, Jiahao and Chen, Hao and Wang, Zhican and Zhou, Hongyu and Zhao, Wanru and Lane, Nicholas and Fan, Hongxiang},
  booktitle={Proceedings of the 30th Asia and South Pacific Design Automation Conference},
  pages={988--994},
  year={2025}
}

@article{hu2024uvllm,
  title={Uvllm: An automated universal rtl verification framework using llms},
  author={Hu, Yuchen and Ye, Junhao and Xu, Ke and Sun, Jialin and Zhang, Shiyue and Jiao, Xinyao and Pan, Dingrong and Zhou, Jie and Wang, Ning and Shan, Weiwei and others},
  journal={arXiv preprint arXiv:2411.16238},
  year={2024}
}

@article{he2025large,
  title={Large language models for eda: Future or mirage?},
  author={He, Zhuolun and Pu, Yuan and Wu, Haoyuan and Qiu, Tairu and Yu, Bei},
  journal={ACM Transactions on Design Automation of Electronic Systems},
  volume={30},
  number={6},
  pages={1--53},
  year={2025},
  publisher={ACM New York, NY}
}

@inproceedings{liu2023verilogeval,
  title={Verilogeval: Evaluating large language models for verilog code generation},
  author={Liu, Mingjie and Pinckney, Nathaniel and Khailany, Brucek and Ren, Haoxing},
  booktitle={2023 IEEE/ACM International Conference on Computer Aided Design (ICCAD)},
  pages={1--8},
  year={2023},
  organization={IEEE}
}

@article{pinckney2025revisiting,
  title={Revisiting verilogeval: A year of improvements in large-language models for hardware code generation},
  author={Pinckney, Nathaniel and Batten, Christopher and Liu, Mingjie and Ren, Haoxing and Khailany, Brucek},
  journal={ACM Transactions on Design Automation of Electronic Systems},
  year={2025},
  publisher={ACM New York, NY}
}

@inproceedings{lu2024rtllm,
  title={Rtllm: An open-source benchmark for design rtl generation with large language model},
  author={Lu, Yao and Liu, Shang and Zhang, Qijun and Xie, Zhiyao},
  booktitle={2024 29th Asia and South Pacific Design Automation Conference (ASP-DAC)},
  pages={722--727},
  year={2024},
  organization={IEEE}
}

@inproceedings{liu2024openllm,
  title={OpenLLM-RTL: Open Dataset and Benchmark for LLM-Aided Design RTL Generation(Invited)},
  author={Liu, Shang and Lu, Yao and Fang, Wenji and Li, Mengming and Xie, Zhiyao},
  booktitle={Proceedings of 2024 IEEE/ACM International Conference on Computer-Aided Design (ICCAD)},
  year={2024},
  organization={ACM}
}

@inproceedings{zhao2025mage,
  title={Mage: A multi-agent engine for automated rtl code generation},
  author={Zhao, Yujie and Zhang, Hejia and Huang, Hanxian and Yu, Zhongming and Zhao, Jishen},
  booktitle={2025 62nd ACM/IEEE Design Automation Conference (DAC)},
  pages={1--7},
  year={2025},
  organization={IEEE}
}

@article{jin2025realbench,
  title={Realbench: Benchmarking verilog generation models with real-world ip designs},
  author={Jin, Pengwei and Huang, Di and Li, Chongxiao and Cheng, Shuyao and Zhao, Yang and Zheng, Xinyao and Zhu, Jiaguo and Xing, Shuyi and Dou, Bohan and Zhang, Rui and others},
  journal={arXiv preprint arXiv:2507.16200},
  year={2025}
}

@article{jiang2020pymtl3,
  title={Pymtl3: A python framework for open-source hardware modeling, generation, simulation, and verification},
  author={Jiang, Shunning and Pan, Peitian and Ou, Yanghui and Batten, Christopher},
  journal={IEEE Micro},
  volume={40},
  number={4},
  pages={58--66},
  year={2020},
  publisher={IEEE}
}

@article{zhao2025pro,
  title={PRO-V: An Efficient Program Generation Multi-Agent System for Automatic RTL Verification},
  author={Zhao, Yujie and Wu, Zhijing and Zhang, Hejia and Yu, Zhongming and Ni, Wentao and Ho, Chia-Tung and Ren, Haoxing and Zhao, Jishen},
  journal={arXiv preprint arXiv:2506.12200},
  year={2025}
}

@inproceedings{hung2004reference,
  title={Reference model based RTL verification: An integrated approach},
  author={Hung, William NN and Narasimhan, Naren},
  booktitle={Proceedings. Ninth IEEE International High-Level Design Validation and Test Workshop (IEEE Cat. No. 04EX940)},
  pages={9--13},
  year={2004},
  organization={IEEE}
}

@inproceedings{firouzi2025chipmnd,
  title={ChipMnd: LLMs for Agile Chip Design},
  author={Firouzi, Farshad and Pan, David Z and Gu, Jiaqi and Farahani, Bahar and Chaudhuri, Jayeeta and Yin, Ziang and Ma, Pingchuan and Domanski, Peter and Chakrabarty, Krishnendu},
  booktitle={2025 IEEE 43rd VLSI Test Symposium (VTS)},
  pages={1--10},
  year={2025},
  organization={IEEE}
}

@inproceedings{ho2025verilogcoder,
  title={Verilogcoder: Autonomous verilog coding agents with graph-based planning and abstract syntax tree (ast)-based waveform tracing tool},
  author={Ho, Chia-Tung and Ren, Haoxing and Khailany, Brucek},
  booktitle={Proceedings of the AAAI Conference on Artificial Intelligence},
  volume={39},
  number={1},
  pages={300--307},
  year={2025}
}

@inproceedings{yu2025spec2rtl,
  title={Spec2RTL-Agent: Automated Hardware Code Generation from Complex Specifications Using LLM Agent Systems},
  author={Yu, Zhongzhi and Liu, Mingjie and Zimmer, Michael and Celine, Yingyan and Liu, Yong and Ren, Haoxing},
  booktitle={2025 IEEE International Conference on LLM-Aided Design (ICLAD)},
  pages={37--43},
  year={2025},
  organization={IEEE}
}

@misc{verilator,
  title        = {Verilator: Open-Source SystemVerilog Simulator and Lint Tool},
  author       = {Snyder, Wilson and Verilator contributors},
  year = {2025},
  howpublished = {\url{https://github.com/verilator/verilator}},
  note = {Released: 2025-12-11}
}

@misc{openai2025gpt52,
  author = {OpenAI},
  title = {Introducing {GPT}-5.2: Professional Knowledge Work and Hallucination Reduction},
  year = {2025},
  howpublished = {\url{https://openai.com/index/introducing-gpt-5-2/}},
  note = {Released: 2025-12-11}
}

@misc{openai2025gpt5,
  author = {OpenAI},
  title = {Introducing {GPT}-5: A Unified System for Reasoning and Multimodality},
  year = {2025},
  howpublished = {\url{https://openai.com/index/introducing-gpt-5/}},
  note = {Released: 2025-08-07}
}

@misc{openai2024gpt4o,
  author = {OpenAI},
  title = {{GPT}-4o: Enhanced Reasoning and Multilingual Performance (2024-11-20 Update)},
  year = {2024},
  howpublished = {\url{https://openai.com/index/gpt-4o-2024-11-20/}},
  note = {Released: 2024-11-20}
}

@misc{openai2024gpt35,
  author = {OpenAI},
  title = {OpenAI Unveils New Models and Lower Price (gpt-3.5-turbo-0125)},
  year = {2024},
  howpublished = {\url{https://openai.com/blog/new-embedding-models-and-api-updates}},
  note = {Released: 2024-01-25}
}

@misc{anthropic2025opus,
  author = {Anthropic},
  title = {Introducing {C}laude {O}pus 4.5: Our Most Intelligent Model},
  year = {2025},
  howpublished = {\url{https://www.anthropic.com/news/claude-opus-4-5}},
  note = {Released: 2025-11-24}
}

@misc{anthropic2025haiku,
  author = {Anthropic},
  title = {Introducing {C}laude {H}aiku 4.5: Speed and Intelligence at Scale},
  year = {2025},
  howpublished = {\url{https://www.anthropic.com/news/claude-haiku-4-5}},
  note = {Released: 2025-10-15}
}

@misc{anthropic2025sonnet,
  author = {Anthropic},
  title = {Introducing {C}laude {S}onnet 4.5: Setting New Records in Reasoning},
  year = {2025},
  howpublished = {\url{https://www.anthropic.com/news/claude-sonnet-4-5}},
  note = {Released: 2025-09-29}
}

@misc{google2025gemini3flash,
  author = {Google DeepMind},
  title = {{G}emini 3 {F}lash: Frontier Intelligence Built for Speed},
  year = {2025},
  howpublished = {\url{https://blog.google/products-and-platforms/products/gemini/gemini-3-flash/}},
  note = {Released: 2025-12-17}
}

@misc{google2025gemini25,
  author = {Google DeepMind},
  title = {{G}emini 2.5 {P}ro and Thinking: The Stable Release},
  year = {2025},
  howpublished = {\url{https://blog.google/technology/ai/gemini-2-5-pro-release/}},
  note = {Released: 2025-06-17}
}

@misc{deepseek2025r1,
  author = {DeepSeek-AI},
  title = {{DeepSeek-R1}: Incentivizing Reasoning Capability in {LLM}s via Reinforcement Learning},
  year = {2025},
  howpublished = {arXiv preprint arXiv:2501.12948},
  note = {Released: 2025-01-20}
}

@misc{deepseek2024v25,
  author = {DeepSeek-AI},
  title = {{DeepSeek-V2.5} Model Upgrade: Merging Chat and Coder Capabilities},
  year = {2024},
  howpublished = {\url{https://api-docs.deepseek.com/updates}},
  note = {Released: 2024-09-05}
}

@misc{meta2024llama33,
  author = {Meta},
  title = {{Llama} 3.3: 70{B} Model Delivering 405{B}-Level Quality},
  year = {2024},
  howpublished = {\url{https://ai.meta.com/blog/llama-3-3-70b/}},
  note = {Released: 2024-12-06}
}

@misc{meta2024llama31,
  author = {Meta},
  title = {The {Llama} 3.1 Collection: 8{B}, 70{B}, and 405{B}},
  year = {2024},
  howpublished = {\url{https://ai.meta.com/blog/llama-3-1/}},
  note = {Released: 2024-07-23}
}

@inproceedings{panda2001systemc,
  title={SystemC: a modeling platform supporting multiple design abstractions},
  author={Panda, Preeti Ranjan},
  booktitle={Proceedings of the 14th international symposium on Systems synthesis},
  pages={75--80},
  year={2001}
}

@inproceedings{wolf2013yosys,
  title={Yosys-a free Verilog synthesis suite},
  author={Wolf, Clifford and Glaser, Johann and Kepler, Johannes},
  booktitle={Proceedings of the 21st Austrian Workshop on Microelectronics (Austrochip)},
  volume={97},
  year={2013}
}

@article{foster20222022,
  title={The 2022 wilson research group functional verification study},
  author={Foster, Harry and others},
  journal={Siemens. com},
  year={2022}
}

@article{keutzer2000system,
  title={System-level design: Orthogonalization of concerns and platform-based design},
  author={Keutzer, Kurt and Newton, A Richard and Rabaey, Jan M and Sangiovanni-Vincentelli, Alberto},
  journal={IEEE transactions on computer-aided design of integrated circuits and systems},
  volume={19},
  number={12},
  pages={1523--1543},
  year={2000},
  publisher={IEEE}
}

@article{price1995pentium,
  title={Pentium FDIV flaw-lessons learned},
  author={Price, Dick},
  journal={IEEE Micro},
  volume={15},
  number={2},
  pages={86--88},
  year={1995},
  publisher={IEEE}
}
\bibliographystyle{icml2026}

\newpage
\appendix
\onecolumn
\section{Appendix}
\subsection{Detailed Model Version Information}
Please refer to ~\autoref{tab:model_details} for model versions we used for evaluation.
We also list the release date, model size, and API names.
\begin{table*}[h]
\centering
\small 
\renewcommand{\arraystretch}{1.1} 

\caption{Overview of Evaluated Models with Verified Release Dates and Parameter Sizes. Model references cite technical reports; sizes with (*) denote industry estimates based on performance benchmarks.}
\label{tab:model_details}

\begin{tabularx}{\textwidth}{@{} l l l l X @{}}
\toprule
\textbf{Model} & \textbf{Reference} & \textbf{Release Date} & \textbf{Size} & \textbf{Model API Name} \\
\midrule

GPT-5.2 & \cite{openai2025gpt52} & 2025-12-11 & 52.5T* & \url{gpt-5.2-2025-12-11} \\
GPT-5   & \cite{openai2025gpt5}  & 2025-08-07 & 52.5T* & \url{gpt-5-2025-08-07} \\
GPT-4o  & \cite{openai2024gpt4o} & 2024-11-20 & $\sim$200B* & \url{gpt-4o-2024-11-20} \\
GPT-3.5 Turbo & \cite{openai2024gpt35} & 2024-01-25 & 175B* & \url{gpt-3.5-turbo-0125} \\

\midrule
Claude 4.5 Opus & \cite{anthropic2025opus}   & 2025-11-24 & $>$150B* & \url{claude-opus-4-5-20251101} \\
Claude 4.5 Haiku & \cite{anthropic2025haiku}  & 2025-10-15 & $<$40B* & \url{claude-haiku-4-5-20251001} \\
Claude 4.5 Sonnet & \cite{anthropic2025sonnet} & 2025-09-29 & 150B* & \url{claude-sonnet-4-5-20250929} \\

\midrule
Gemini 3 Flash & \cite{google2025gemini3flash}   & 2025-12-17 & $\sim$10B* & \url{gemini-3-flash-preview} \\
Gemini 2.5 Pro & \cite{google2025gemini25}  & 2025-06-17 & 400B* & \url{gemini-2.5-pro} \\

\midrule
DeepSeek-R1 & \cite{deepseek2025r1}          & 2025-01-20 & 671B & \url{deepseek-reasoner} \\
DeepSeek-Coder & \cite{deepseek2024v25} & 2024-09-05 & 236B & \url{deepseek-coder} \\

\midrule
Llama 3.3 70B & \cite{meta2024llama33} & 2024-12-06 & 70B & \url{Llama-3.3-70B-Instruct-Turbo} \\
Llama 3.1 8B & \cite{meta2024llama31}  & 2024-07-23 & 8B & \url{Meta-Llama-3.1-8B-Instruct-Turbo} \\

\bottomrule
\end{tabularx}
\end{table*}

\subsection{Detailed Information of Modules}
We summarized the description of each model in ~\autoref{tab:module_descriptions_part1}.

\begin{table*}[t]
\centering
\caption{Verilog Module Descriptions}
\label{tab:module_descriptions_part1}
\begin{small} 
\begin{tabular}{p{0.07\textwidth} p{0.31\textwidth} p{0.3\textwidth} r r}
\toprule
\textbf{Design} & \textbf{Module Name} & \textbf{Description} & \textbf{Code Lines} & \textbf{Circuit Cells}\\
\midrule
Self-Contained 
 & continuous\_input\_sequence\_detect & 8-bit shift-register pattern detector & 33 & 16\\
 & extraneous\_items\_sequence\_detect & 9-bit shift-register detector  & 51 & 16\\
 & non-overlapping\_sequence\_detect & Non-overlapping 6-bit group sequence checker & 95 & 35\\
 & continuous\_input\_sequence\_detect2 & FSM sequence detector  & 78 & 29\\
 & signal\_generator & 5-bit waveform generator & 48 & 112\\
 & data\_serial-to-parallel\_circuit & Serial-to-parallel packer & 50 &28\\
 & data\_accumulation\_output & Ready/valid accumulator & 44 &91\\
 & FracWidthConv\_24to128 & Bit-pack a valid-only 24-bit input stream into 128-bit outputs & 48 &532\\
 & FracWidthConv\_8to12 & Bit-pack a valid-only 8-bit input stream into 12-bit outputs & 48 &42\\
 & IntWidthConv\_8to16 & Valid-only width converter that packs two 8-bit inputs into one 16-bit output & 40 &28\\
 & 4-bit\_carry\_look-ahead\_adder\_circuit & 4-bit carry look-ahead adder & 18 &20\\
 & state\_machine\_overlap\_sequence\_detection & Overlapping sequence detector & 40 &19\\
 & least\_common\_multiple & 8-bit LCM/GCD calculator & 92 &7409\\
 & sequence\_generator & Cyclic sequence generator & 21 &7\\
 & vending\_machine2 & Vending machine FSM & 64 & 92\\
 & OddDiv\_HalfDuty & Divide-by-7 clock divider with $\sim$50\% duty cycle & 56 &22\\
 & arbitrary\_fractional\_frequency\_division & Fractional clock divider achieving average /8.7 & 51 &66\\
 & write\_state\_machine\_two\_stage & Two-process Moore FSM & 53&11 \\
 & gray\_code\_counter & 4-bit Gray-code counter & 44 & 25\\
 & multiplication\_and\_bitwise\_operations & Resource-minimized constant multiply compute & 12 &96\\
 & pulse\_synchronization\_circuit & Clock-domain crossing pulse synchronizer & 45 &6\\
 & simple\_stopwatch & Stopwatch counter & 38 & 48\\
 & settable\_counter & Hex counter & 50 &22\\
 & up\_and\_down\_counter & Decimal up/down counter & 44 &33\\
 & simple\_implementation\_RAM & Simple dual-port RAM (depth 8, width 4) & 31 &189\\
 & johnson\_counter & 4-bit Johnson counter & 12 &5\\
 & pipeline\_multiplier & 4-bit unsigned pipelined multiplier & 42 &93\\
 & traffic\_lights & Traffic light controller & 101 &105\\
 & gaming\_machine\_billing\_program & Prepaid gaming billing FSM & 36 & 178\\
\midrule
Not-Self-Contained 
 & submodules\_to\_comparison\_of\_three\_inputs & A Comparator for the minimum of three 8-bit unsigned inputs & 56 & 116 \\
 & full\_subtractor\_using\_three\_to\_eight\_decoder & A 1-bit full subtractor & 109 & 28\\
 & asynchronous\_FIFO & Asynchronous FIFO (dual clock) & 144 & 380\\
 & synchronous\_FIFO & Single-clock FIFO & 88 & 367\\
 & 8\_bit\_alu & A 8-bit ALU & 70 & 81\\
 & cpu\_top & A Top-level 16-bit CPU & 250 & 1195\\
\midrule
cpu-design 
 & controller & A RISC-V main controller that decodes the 7-bit opcode into datapath control signals & 52 & 42\\
 & alu & Purely combinational 32-bit ALU with 4-bit `Operation` code & 51 & 1456\\
 & regfile & 32×32-bit register file with 1 write port and 2 combinational read ports & 39 & 3122\\
 & branch\_unit & A branch unit & 26 & 206\\
 & alu\_controller & A RISC-V ALU-control decoder & 45 & 69\\
 & pc\_reg & A PC register & 45 & 180\\
 & div & A Sequential 32-bit signed/unsigned division unit & 99 & 1228\\
 & ctrl & A Pipeline control Unit & 73 & 213\\
 & cp0\_reg & A CP0 register file implementing Count/Compare/Status/Cause/EPC/etc. & 183 & 1245\\
\bottomrule
\end{tabular}
\end{small}
\end{table*}

\end{document}